\documentclass[preprint,12pt,authoryear]{elsarticle}

\usepackage{amssymb}
\usepackage{amsmath}
\usepackage{longtable}

\newcommand{\coloneqq}{\mathrel{\mathop:}=}

\journal{AI Open}

\begin{document}

\begin{frontmatter}



\title{Bio-Inspired Adaptive Neurons For Dynamic Weighting In Artificial Neural Networks}


\author[inst1]{Ashhadul Islam}

\affiliation[inst1]{organization={College Of Science \& Engineering, Hamad Bin Khalifa University},
            addressline={Education City}, 
            city={Doha},
            postcode={34110}, 
            country={Qatar}}

\author[inst1,inst2]{Abdesselam Bouzerdoum}
\author[inst1]{Samir Brahim Belhaouari}

\affiliation[inst2]{organization={School of Electrical, Computer \& Telecommunications Engineering, University of Wollongong},
            addressline={Address Two}, 
            city={Wollongong},
            postcode={2500}, 
            state={New South Wales},
            country={Australia}}

\begin{abstract}
Traditional neural networks employ fixed weights during inference, limiting their ability to adapt to changing input conditions, unlike biological neurons that adjust signal strength dynamically based on stimuli. This discrepancy between artificial and biological neurons constrains neural network flexibility and adaptability. To bridge this gap, we propose a novel framework for adaptive neural networks, where neuron weights are modeled as functions of the input signal, allowing the network to adjust dynamically in real-time. Importantly, we achieve this within the same traditional architecture of an Artificial Neural Network, maintaining structural familiarity while introducing dynamic adaptability. In our research, we apply Chebyshev polynomials as one of the many possible decomposition methods to achieve this adaptive weighting mechanism, with polynomial coefficients learned during training.  Out of the 145 datasets tested, our adaptive Chebyshev neural network demonstrated a marked improvement over an equivalent MLP in approximately 83\% of cases, performing strictly better on 121 datasets. In the remaining 24 datasets, the performance of our algorithm matched that of the MLP, highlighting its ability to generalize standard neural network behavior while offering enhanced adaptability. As a generalized form of the MLP, this model seamlessly retains MLP performance where needed while extending its capabilities to achieve superior accuracy across a wide range of complex tasks. These results underscore the potential of adaptive neurons to enhance generalization, flexibility, and robustness in neural networks, particularly in applications with dynamic or non-linear data dependencies.
\end{abstract}

\begin{graphicalabstract}
\includegraphics[width=0.95\textwidth]{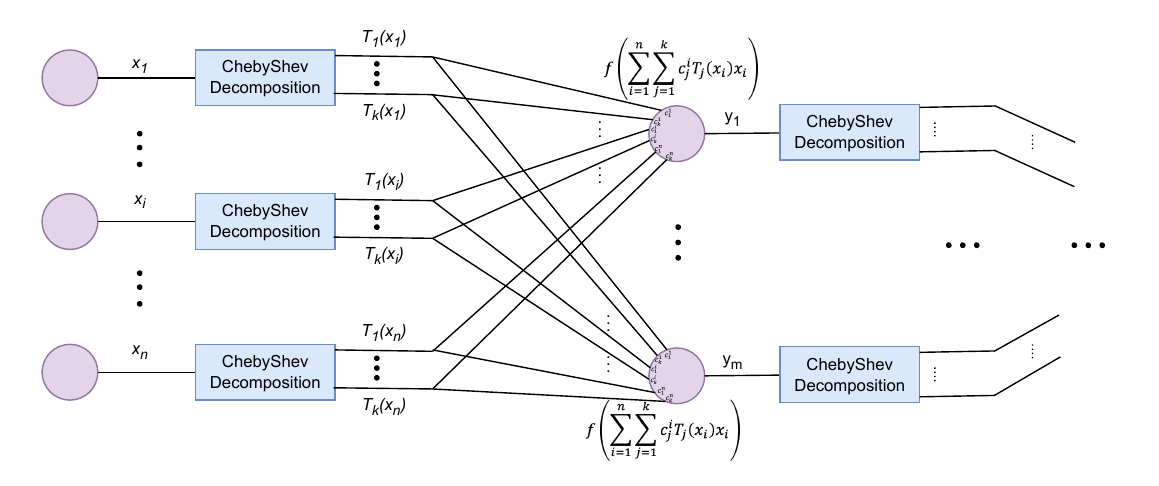}
\end{graphicalabstract}

\begin{highlights}
\item Dynamic weighting mechanism allows neurons to adapt their output based on input strength, mimicking biological adaptability.  
\item Chebyshev polynomials provide mathematical flexibility, enabling nuanced and complex weight dynamics.  
\item Improved generalization capability in tasks involving complex and non-linear input-output relationships.  
\end{highlights}

\begin{keyword}
Adaptive neural networks \sep Dynamic weighting \sep Input-dependent weights \sep Neural network generalization \sep Chebyshev polynomials
\PACS 0000 \sep 1111
\MSC 0000 \sep 1111
\end{keyword}

\end{frontmatter}



\section{Introduction}
In conventional neural networks \citep{bishop1994neural}, the weights associated with each neuron remain constant during inference, meaning that the output is determined by a fixed set of weights that were learned during training. While this method has been successful in numerous applications, it does not mirror the adaptive behavior observed in biological neurons. In biological systems, neurons adjust the strength of the signals they transmit in response to the intensity of the input stimuli, leading to dynamic weighting \citep{rohe2021inputs,yang2024co,brito2024learning}. This discrepancy between artificial neurons and their biological counterparts may limit the flexibility and adaptability of current neural networks, particularly in environments with variable input conditions.

To address this, we propose a bio-inspired approach that incorporates adaptive neurons using Chebyshev polynomials \citep{mason2002chebyshev}. In this framework, instead of having fixed weights, the weight of each connection dynamically adjusts based on the input signal. Specifically, the weight is expressed as a sum of Chebyshev polynomials, where the coefficients of the polynomials are learnable parameters optimized during training.

Let \( x_i \) represent the input to the neuron and \( w_i \) represent the adaptive weight associated with the input \( x_i \). The adaptive weight is defined as:

\begin{equation}
w_i(x_i) = \sum_{j=0}^{k} c_{i,j} T_j(x_i)    
\end{equation}

where \( c_{i,j} \) are the learnable coefficients, \( T_j(x_i) \) represents the Chebyshev polynomial of the first kind of order \( j \) evaluated at \( x_i \), and \( k \) is the polynomial order. The neuron’s output is then computed as:

\begin{equation}
\text{Neuron Output: } y = \sum_{i=1}^{n} x_i \cdot w_i (x_i) = \sum_{i=1}^{n} x_i \cdot \sum_{j=0}^{k} c_{i,j} T_j(x_i)
\end{equation}

By making the weights dependent on the input, the neuron can dynamically adjust its output, allowing it to better mimic the adaptive behavior seen in biological neurons. This results in a more flexible and biologically plausible artificial neuron model.

This approach introduces several key innovations: 
\begin{enumerate}
    \item Dynamic weighting, enabling neuron output to reflect the input strength in a biologically plausible manner
    \item Mathematical flexibility provided by Chebyshev polynomials, enabling nuanced and complex weight dynamics
    \item Enhanced generalization in tasks with complex or non-linear input-output relationships.
\end{enumerate}



Traditional neural networks employ fixed weights that remain static after training, regardless of input variations. This approach can limit the model's adaptability across varying conditions. In contrast, our proposed framework introduces adaptive weights, where each weight dynamically changes as a function of the input data. By leveraging decomposition functions such as Chebyshev polynomials, the model represents each weight as a sum of polynomial terms, enabling the network to flexibly adjust to a wide range of input patterns. This design enhances the model’s responsiveness and draws inspiration from the behavior of biological neurons, which adaptively modulate their responses based on stimuli.

Our method bridges the gap between biological and artificial neurons, improving adaptability and robustness in neural networks, with potential applications in areas like bio-inspired computing, brain-computer interfaces, and pattern recognition. This research paves the way for new architectures that incorporate dynamic weight adjustment mechanisms, advancing the field of neural network design.

\subsection{Biological Basis of Neurons and Their Inspiration for Artificial Neural Networks}

\subsubsection{Human Neuron Structure and Function}

\begin{figure}[htbp]
    \centering
    \includegraphics[width=0.6\textwidth]{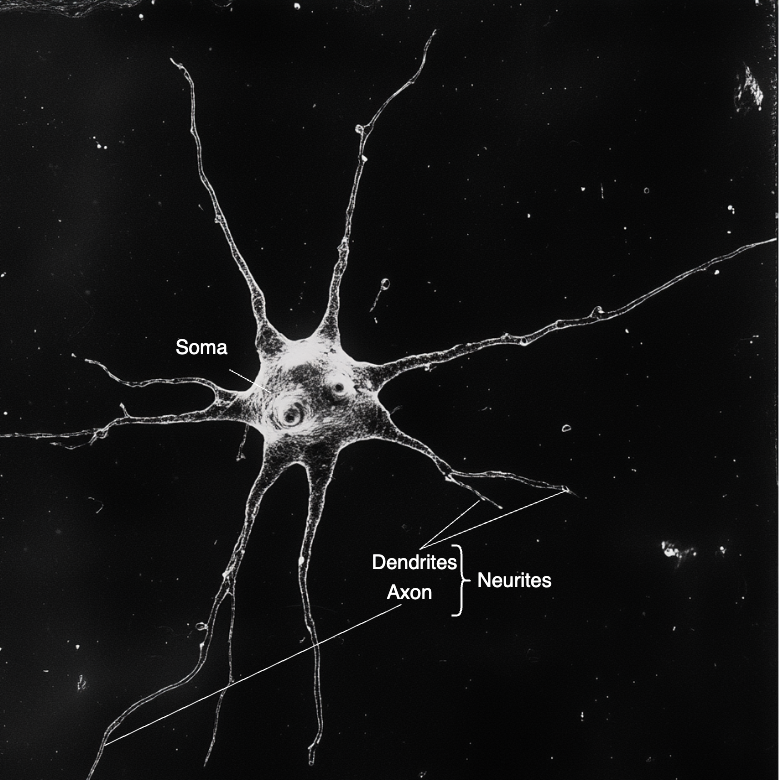}
    \caption{Structure of a human neuron, depicting the soma (cell body), dendrites, and axon, which are essential for transmitting neural signals.}
    \label{fig:neuron}
\end{figure}

Human neurons are specialized cells responsible for transmitting signals within the brain and nervous system, enabling cognition, sensation, and motor control. As shown in Figure \ref{fig:neuron}, a neuron consists of three primary components: the soma, dendrites, and axon \citep{kandel2000principles}. The soma, or cell body, contains the nucleus and is responsible for maintaining the neuron's structure and performing essential metabolic processes \citep{alberts2002molecular}. Dendrites are branch-like structures that receive chemical signals (neurotransmitters) from other neurons and convert these chemical signals into electrical impulses that travel toward the soma \citep{spruston1999dendritic}. The axon is a long projection that transmits these electrical signals away from the soma, culminating at the synapse, where neurotransmitters are released to communicate with other neurons \citep{hoy2016principles}.

\subsubsection{Neuron Communication and Nonlinearity}
Neurons communicate via electrical and chemical signals, enabling rapid and complex information processing across the brain \citep{bear2007exploring}. When a neuron is activated, it generates an action potential—an electrical impulse initiated when the neuron's membrane depolarizes due to the influx of sodium ions (Na+), allowing the neuron to reach a critical threshold \citep{hille2001ion}. This action potential travels along the axon and is converted into a chemical signal at the synapse, where neurotransmitters are released and bind to receptors on the dendrites of a postsynaptic neuron, generating an electrical response in the receiving neuron \citep{kandel2000principles}.

The response of neurons to input signals is inherently nonlinear. Neurons exhibit threshold behavior, where they fire action potentials only when the input surpasses a certain threshold \citep{koch2004biophysics}. Below this threshold, the neuron remains inactive, regardless of slight variations in input. Furthermore, neurons integrate signals over time through temporal summation, where multiple inputs arriving within a short time window can summate, potentially pushing the neuron past its threshold. Spatial summation also occurs, where inputs from different locations on the dendrites have varying effects on the soma, depending on their proximity and strength \citep{bliss1993synaptic}. Another important factor contributing to nonlinearity is synaptic plasticity, where the strength of synaptic connections changes based on prior activity, thus modifying how neurons communicate over time.

\subsubsection{Inspiration for Artificial Neural Networks}
Artificial neural networks (ANNs) are inspired by biological neurons but simplify their functions to make them computationally efficient \citep{schmidhuber2015deep}. While artificial neurons process input via linear combinations of weights and biases, biological neurons exhibit more dynamic and nonlinear behaviors. Understanding these nonlinear properties in biological neurons offers valuable insights for improving ANN design \citep{koch2000role}.

In biological neurons, synaptic plasticity allows connections to strengthen or weaken over time, adapting to stimuli dynamically \citep{bliss1993synaptic}. Similarly, ANNs adjust synaptic weights during training to minimize error, but without the temporal and spatial complexity of biological learning. While activation functions in ANNs (such as ReLU or sigmoid) introduce nonlinearity, they are a simplified abstraction of the more complex processes that govern biological neuron responses \citep{glorot2011deep}.

Recent advancements such as spiking neural networks (SNNs), which more closely model biological neurons by simulating spike-timing dynamics, introduce temporal nonlinearity into the network \citep{pfeiffer2018deep}. Hebbian learning, based on the principle that "cells that fire together, wire together," is another biologically inspired mechanism that could improve how ANNs adapt to changing environments \citep{hebb2005organization}.  Models such as recurrent neural networks (RNNs) and Long Short-Term Memory (LSTM) networks are used to capture temporal dependencies, mimicking the recurrent connections in biological circuits \citep{hochreiter1997long}. These architectures bring ANNs closer to the dynamic, feedback-rich environment of biological neurons.

\subsubsection{Relevance to Adaptive Neural Networks}
The nonlinear, dynamic behavior of biological neurons, particularly through mechanisms like synaptic plasticity and action potentials, has inspired our approach to adaptive neurons. Biological neurons adjust their responses based on the input signal's intensity and timing, a process we aim to emulate in artificial neurons using Chebyshev polynomials. By incorporating input-dependent adaptive weights, we propose a model that mirrors the biological neuron's ability to integrate complex signals nonlinearly. This provides the flexibility needed to capture intricate patterns in data, much like how biological systems process information.

\subsection{Orthogonal Decomposition Techniques and Their Application to Adaptive Neurons}
In various mathematical and computational problems, approximating a function \( f(x) \) using orthogonal basis functions is a widely used method for capturing the complexity of a function while ensuring computational efficiency. This technique, known as \textbf{orthogonal decomposition} \citep{boyd2001chebyshev}, involves representing a function as a sum of orthogonal basis functions \( \{ \phi_n(x) \} \), where each term in the series contributes uniquely to different features of the function \citep{press2007numerical}. The general form of the approximation is given by:

\[
f_N(x) = \sum_{n=0}^{N} c_n \phi_n(x)
\]

where \( \phi_n(x) \) are orthogonal basis functions, and \( c_n \) are coefficients determined by projecting \( f(x) \) onto the basis functions. \(N\) represents the degree of the approximation or the highest order of the orthogonal basis functions used in the summation

\subsection{Overview of Orthogonal Decomposition Techniques}

Orthogonal decomposition is a foundational technique in functional approximation, where a function \( f(x) \) is expressed as a weighted sum of orthogonal basis functions:
\[
f_N(x) = \sum_{n=0}^{N} c_n \phi_n(x)
\]
where \( \phi_n(x) \) are orthogonal functions and \( c_n \) are projection coefficients.

Among various orthogonal bases, several notable examples include:

\begin{itemize}
    \item \textbf{Legendre Polynomials}: Commonly used over the interval \( [-1, 1] \), with orthogonality under a uniform weight function.

    \item \textbf{Wavelets}: Useful when both spatial and frequency localization are needed, though typically more computationally intensive.

    \item \textbf{Eigenfunction Expansions}: Effective in representing solutions to operator-based equations but less practical for efficient neural computations.

    \item \textbf{Chebyshev Polynomials}: Particularly advantageous for neural network integration due to their minimax error properties, orthogonality with respect to a weight function \( \frac{1}{\sqrt{1 - x^2}} \), and efficient recursive computation.
\end{itemize}

In this work, we specifically choose Chebyshev polynomials as the basis for adaptive weight modeling, as they offer a compelling balance of approximation power, computational efficiency, and numerical stability, which are essential in neural network training and deployment.

\subsection{Adaptive Weighting through Chebyshev Polynomial Decomposition}

In the context of our proposed adaptive neuron model, we leverage  \textbf{Chebyshev polynomials} for orthogonal decomposition of the neuron weights. By representing the weight as a sum of Chebyshev polynomials, we introduce dynamic, input-dependent behavior that enables the neuron to adjust its output based on the input signal. The adaptive weight is expressed as:

\[
w_i(x_i) = \sum_{j=0}^{k} c_{i,j} T_j(x_i)
\]

This method allows the network to dynamically adjust its weights based on the input, leading to better flexibility and generalization, particularly in tasks where the input-output relationship is non-linear or complex.

By incorporating orthogonal decomposition into the design of the neuron model, we ensure that the learned weights are mathematically well-behaved and capable of capturing diverse patterns in the input data. Chebyshev polynomials, with their orthogonality and efficient approximation properties, play a central role in making this approach both computationally feasible and effective.

\subsection{Properties of Chebyshev Polynomials}

Chebyshev polynomials are a sequence of orthogonal polynomials that arise in various fields of applied mathematics, including approximation theory and numerical analysis. They are particularly useful in scenarios that involve minimizing the maximum error between a function and its polynomial approximation, making them well-suited for tasks requiring efficient and accurate representations of complex functions. The key properties of Chebyshev polynomials, particularly those of the first kind \( T_j(x) \), which are used in our adaptive neuron model \citep{mason2002chebyshev}, include the following:

\subsubsection{Recursive Definition}
Chebyshev polynomials of the first kind, denoted \( T_j(x) \), can be defined recursively. The first two polynomials in the sequence are:

\[
T_0(x) = 1
\]
\[
T_1(x) = x
\]

For \( j \geq 2 \), the polynomials follow the recursive relation:

\[
T_j(x) = 2x \cdot T_{j-1}(x) - T_{j-2}(x)
\]

This recursive definition \citep{mason2002chebyshev} allows for efficient computation of higher-order polynomials and facilitates their integration into the adaptive neuron framework without significantly increasing computational overhead.

\subsubsection{Orthogonality}
One of the key features of Chebyshev polynomials is their orthogonality with respect to the weight function \( \frac{1}{\sqrt{1 - x^2}} \) over the interval \( [-1, 1] \) \citep{mason2002chebyshev}. Specifically, for \( j \neq m \):

\[
\int_{-1}^{1} T_j(x) T_m(x) \frac{dx}{\sqrt{1 - x^2}} = 0
\]

Orthogonality ensures that the polynomials represent independent components of the input, which can lead to better representation and learning of complex, high-dimensional input data in neural networks. In the context of our model, the orthogonality property aids in reducing redundancy in the neuron’s response to varying inputs \citep{mason2002chebyshev}.

\subsubsection{Extremal Properties}
Chebyshev polynomials are known for minimizing the maximum deviation from zero over the interval \( [-1, 1] \). This extremal property makes them particularly effective in approximating functions with high accuracy while keeping the coefficients well-behaved. This feature is highly desirable in neural networks, as it can help prevent overfitting by providing smooth, well-distributed weights across inputs \citep{mason2002chebyshev,trefethen2009approximation,rivlin2020chebyshev}.

\subsubsection{Equioscillation and Roots}
The roots of the Chebyshev polynomials of the first kind, \( T_j(x) \), are located at:

\[
x_k = \cos\left(\frac{(2k - 1) \pi}{2j}\right) \quad \text{for} \quad k = 1, 2, \ldots, j
\]

These roots are distributed symmetrically in the interval \( [-1, 1] \) and are used in approximation theory to achieve optimal interpolation. In our adaptive neuron model, the roots of the Chebyshev polynomials provide a natural way to discretize and sample the input space, allowing for dynamic adjustment of weights based on input intensities \citep{mason2002chebyshev,trefethen2009approximation,rivlin2020chebyshev}.

\subsubsection{Rapid Growth Outside of \( [-1, 1] \)}
While Chebyshev polynomials are well-behaved within the interval \( [-1, 1] \), their values grow rapidly for inputs outside this interval. This behavior can be controlled within our model by normalizing the inputs or using a transformation that ensures the input remains in the range where the polynomials are stable. This is critical in ensuring that the adaptive weights do not become overly sensitive to out-of-range inputs, thereby preserving the stability and robustness of the neuron’s output \citep{boyd2001chebyshev,mason2002chebyshev,trefethen2009approximation,rivlin2020chebyshev}.

\subsubsection{Approximation Power}
Chebyshev polynomials are widely recognized for their superior approximation properties, especially when compared to other polynomials such as Legendre or Laguerre. The series expansion in terms of Chebyshev polynomials is often used in Chebyshev approximation to achieve a near-optimal polynomial approximation of continuous functions. In our model, this approximation power translates into the ability to represent complex, non-linear relationships between the input and the neuron’s response, enabling more nuanced and flexible learning \citep{mason2002chebyshev,trefethen2009approximation,rivlin2020chebyshev}.

\subsection{Chebyshev Polynomials in Adaptive Neural Networks}

The properties of Chebyshev polynomials make them an excellent choice for adaptive neural networks. Their orthogonality, extremal properties, and efficient recursive computation allow neurons to dynamically adjust their weights based on input signals, enhancing the network’s ability to generalize and adapt to diverse input conditions. The integration of Chebyshev polynomials in the weight computation, as proposed in this research, leads to several advantages:

\begin{itemize}
    \item \textbf{Dynamic and Adaptive Weighting:} Unlike traditional neurons with fixed weights, neurons in our model can adjust their weights on-the-fly based on the intensity of the input, providing more flexibility and adaptability to changing environments.
    
    \item \textbf{Improved Generalization:} The orthogonality and approximation power of Chebyshev polynomials help in reducing overfitting and improving generalization to unseen data, especially in tasks with complex input-output relationships.
    
    \item \textbf{Efficient Computation:} The recursive formulation of Chebyshev polynomials ensures that the additional computational overhead introduced by dynamic weighting remains manageable, making the approach feasible for real-world applications.
\end{itemize}

\subsection{Rationale of Choosing Chebyshev Polynomials}

Among the many available families of orthogonal basis functions, we selected Chebyshev polynomials for their unique combination of mathematical and practical properties that align well with the goals of our adaptive neuron framework. Specifically, Chebyshev polynomials:

\begin{itemize}
    \item Exhibit \textbf{minimax approximation properties}, meaning they minimize the maximum error between the true function and its approximation. This is particularly advantageous in neural networks, where it can help prevent large deviations and promote stable learning.
    
    \item Possess strong \textbf{orthogonality characteristics} under the weighted inner product space with weight function \( \frac{1}{\sqrt{1 - x^2}} \), enabling independent and efficient function decomposition. This reduces redundancy and overfitting in weight adaptation.

    \item Are defined with a \textbf{simple and efficient recursive formulation}, which facilitates fast computation of higher-order terms. This makes their integration into neural architectures computationally feasible even as polynomial order increases.

    \item Have roots and extrema distributed in a way that leads to optimal interpolation nodes (Chebyshev nodes), further improving numerical stability and convergence behavior during training.

    \item Are known to perform better than other orthogonal polynomials such as Legendre and Hermite in approximating smooth functions with fewer coefficients due to their \textbf{superior approximation power}.
\end{itemize}

While other orthogonal bases such as Legendre or Hermite polynomials could also be used, Chebyshev polynomials provide a more favorable balance between approximation accuracy, numerical stability, and computational efficiency. These attributes make them especially well-suited for adaptive neural architectures that aim to dynamically adjust weights in response to input stimuli. These properties, along with the dynamic nature of the proposed adaptive neuron model, open up new possibilities for designing neural networks that are more flexible, robust, and capable of handling non-linear and complex data patterns effectively.

\section{Related Work}
This section provides an overview of recent developments in neural network architectures that enhance adaptability and efficiency through various approaches. We discuss methods that improve inference efficiency, introduce dynamic weighting, leverage Chebyshev polynomials, and incorporate spiking mechanisms, all contributing to the evolution of neural networks toward more adaptive and biologically plausible models. Additionally, we explore the Kolmogorov–Arnold Network (KAN) as a foundation for developing adaptive neuron models.

\subsection{Adaptive Neural Networks for Efficient Inference}
Recent advancements in adaptive neural networks have focused on enhancing computational efficiency by modifying network evaluation based on the complexity of individual examples. One notable approach introduces a method to selectively activate network components, allowing early exits for instances that are accurately classified within initial layers, thus avoiding full model evaluation \citep{bolukbasi2017adaptive}. Additionally, this method introduces a mechanism for adaptive network selection, where a lightweight model is chosen for simpler examples, while more complex networks are reserved for challenging cases. By formulating this process as a policy learning task, the approach optimizes layer-wise or model-level selection through weighted binary classification, significantly reducing inference time while maintaining accuracy.

\subsection{Dynamic Weighting in Neural Networks}
Another prominent direction in adaptive neural networks involves dynamically adjusting weights \citep{han2021dynamic} based on input characteristics during inference. This concept is utilized in various forms, such as attention mechanisms, conditionally parameterized convolutions, and deformable convolutions, each catering to specific model adaptation needs.

For instance:
\begin{itemize}
    \item \textbf{CondConv:} Conditionally parameterized convolutions employ customized convolutional filters for each example, thus enhancing network capacity while maintaining efficient inference \citep{yang2019condconv}.
    \item \textbf{Dynamic Convolution:} Here, multiple convolutional kernels are aggregated dynamically using input-dependent attention, allowing the model to maintain a low computational footprint while increasing representational flexibility \citep{chen2020dynamic}.
    \item \textbf{Segmentation-aware CNNs:} These networks use local attention masks that selectively attend to region-specific inputs, improving spatial precision for tasks such as semantic segmentation and optical flow \citep{harley2017segmentation}.
    \item \textbf{Deformable Convolutional Networks (DCNs):} By adjusting   convolutional sampling locations, these networks adapt their receptive fields to object shapes, which improves performance in tasks requiring spatial sensitivity \citep{dai2017deformable}.
\end{itemize}

\subsection{Chebyshev Polynomials in Neural Networks}

The study by Troumbis \citep{troumbis2020chebyshev} introduces a neural network architecture that leverages Chebyshev polynomials for effective modeling of complex, non-linear environmental data. This network architecture applies Chebyshev polynomials through a layered feedforward structure, with the network parameters optimized via differential evolution, a population-based optimization algorithm.

The network is composed of four sequential layers, each performing a distinct operation:

\begin{itemize}
    \item \textbf{Layer 1:} Generates linear combinations of the input variables \( x_i \) to form \( L_i \) as follows:
    \[
    L_i = \sum_{j=1}^{p} a_{i,j} \cdot x_j
    \]
    where \( a_{i,j} \) represents the learned coefficients, and \( p \) denotes the number of input variables.

    \item \textbf{Layer 2:} Scales each \( L_i \) to the interval \([-1, 1]\) to fit within the domain of Chebyshev polynomials, using the following transformation:
    \[
    \tilde{L_i} = \frac{2(L_i - L_{min})}{L_{max} - L_{min}} - 1
    \]
    where \( L_{min} \) and \( L_{max} \) denote the minimum and maximum values, respectively.

    \item \textbf{Layer 3:} Computes truncated Chebyshev series expansions for each normalized \( \tilde{L_i} \), represented as:
    \[
    T_i(\tilde{L_i}) = \sum_{n=0}^{N} c_{i,n} \cdot T_n(\tilde{L_i})
    \]
    where \( c_{i,n} \) are coefficients specific to each input, \( T_n(\tilde{L_i}) \) represents the Chebyshev polynomial of order \( n \), and \( N \) is the truncation order.

    \item \textbf{Layer 4:} Linearly combines the truncated Chebyshev series from each node to produce the final network output:
    \[
    y = \sum_{i=1}^{c} w_i \cdot T_i(\tilde{L_i})
    \]
    where \( w_i \) represents the weights applied to each truncated series, and \( c \) is the number of hidden nodes.

\end{itemize}

The training of this Chebyshev polynomial-based network is accomplished using the Differential Evolution (DE) algorithm, which optimizes the network parameters through mutation, crossover, and selection phases. DE is particularly suited for this application because of its ability to handle non-linear optimization problems without relying on gradient-based methods, thus providing robustness against local minima and ensuring convergence to an optimal solution.

In experiments, this Chebyshev polynomial network outperformed other architectures, including networks using Hermite polynomials, radial basis functions, and Takagi-Sugeno-Kang neuro-fuzzy models, across diverse environmental datasets. This performance was assessed using standard metrics such as Root Mean Square Error (RMSE) and Mean Absolute Error (MAE), demonstrating the network’s competitive edge in complex, data-intensive tasks.


\paragraph{Theoretical Motivation.}  The use of Chebyshev polynomials is theoretically motivated by their minimax approximation property:  among all polynomials of a given degree, Chebyshev polynomials minimize the maximum deviation from the  target function on the interval $[-1,1]$ \citep{trefethen2019approximation,rivlin2020chebyshev}.  This property ensures that the expanded feature space spans an orthogonal basis with near-optimal  approximation guarantees, thereby enhancing the expressivity of the network compared to standard MLPs,  which rely solely on fixed nonlinear activations such as ReLU or tanh. By embedding this basis into the  weight generation process, Chebyshev Adaptive Networks are able to approximate more complex functions  with fewer parameters or layers, offering a principled improvement in representational capacity. 

\subsubsection{Differences from Troumbis et al. (2020)}

While both our work and Troumbis et al. \citep{troumbis2020chebyshev} utilize Chebyshev polynomials within neural architectures, the underlying philosophy, design, and learning mechanisms differ significantly:

\begin{itemize}
    \item \textbf{Static vs. Adaptive Weighting:} Troumbis et al. employ Chebyshev polynomials as part of a fixed series expansion to approximate the network output. In contrast, our method uses Chebyshev polynomials to define input-dependent adaptive weights at the neuron level. This dynamic formulation allows weights to change with varying input stimuli, thereby offering greater biological plausibility and flexibility.

    \item \textbf{Learning Mechanism:} Troumbis et al. use a population-based optimization technique (Differential Evolution) for training, which does not rely on gradient descent. Our model, on the other hand, is trained using standard backpropagation, making it compatible with mainstream deep learning frameworks and scalable to larger datasets and architectures.

    \item \textbf{Architectural Scope and Generality:} The model in Troumbis et al. is tailored to specific regression problems with a fixed architectural structure. Our model generalizes this by integrating polynomial-based adaptive weights directly into the neuron formulation, making it applicable across a broad range of tasks (classification, regression) and easily extensible to other orthogonal decompositions beyond Chebyshev.

    \item \textbf{Biological Motivation and Neural Interpretation:} While Troumbis et al. focus primarily on mathematical function approximation, our formulation is bio-inspired—aiming to model the dynamic response characteristics of biological neurons where synaptic strength adapts in real-time based on the nature of incoming signals.
\end{itemize}

In summary, our approach significantly extends the scope and applicability of Chebyshev-based neural models by introducing dynamic adaptability, backpropagation-based learning, and broader architectural generalization, thereby addressing a different set of scientific and practical challenges.

\subsection{Spiking Neural Networks and Adaptation}

Spiking Neural Networks (SNNs) \citep{guo2023direct} represent a brain-inspired computational model that utilizes binary spike-based communication, allowing for efficient information processing through event-driven and spatio-temporal mechanisms. These properties make SNNs particularly effective in handling temporal data and enable energy-efficient computations, aligning with biological neural networks’ sparse firing behavior. However, the discontinuous nature of spike-based information transmission introduces challenges for training deep SNNs, as conventional gradient-based optimization methods cannot be directly applied due to the non-differentiable spiking mechanism.

One approach to overcome this challenge is \textit{Spike-Timing-Dependent Plasticity} (STDP) \citep{lobov2020spatial}, a biologically inspired mechanism for weight updates in SNNs. Despite its grounding in biology, STDP alone is not sufficient for training large-scale networks, limiting its applicability in practical, complex tasks. 

Two prevalent methods have been developed to achieve more effective training of deep SNNs:
\begin{enumerate}
    \item \textbf{ANN-to-SNN Conversion}: This method involves training an Artificial Neural Network (ANN) with standard continuous activation functions (such as ReLU) and subsequently converting it into an SNN by replacing the activations with spike-based mechanisms \citep{han2020deep}. This process retains the original network's learned representations and is straightforward to implement. However, it is typically constrained to rate-coding, ignoring the potential for more dynamic temporal behaviors within SNNs.
    
    \item \textbf{Surrogate Gradient (SG) Approach}: The SG approach introduces a differentiable surrogate function that approximates the spiking neuron’s non-differentiable firing activity, enabling backpropagation in SNNs \citep{fang2021deep}. This technique has shown significant promise in handling temporal data and achieving competitive performance with few time steps on large-scale datasets. The SG method allows SNNs to utilize the spatio-temporal dynamics of spikes effectively, thereby enhancing the network’s ability to process temporal information while maintaining efficiency.
\end{enumerate}

In recent studies, including Wu et al. \citep{wu2019direct}, various enhancements to direct learning-based SNNs are discussed, categorized broadly into accuracy improvement methods, efficiency improvement methods, and methods that exploit temporal dynamics. These categories are further divided based on specific objectives:
\begin{itemize}
    \item \textit{Accuracy Improvement}: Focused on increasing representational capacity and alleviating training challenges \citep{wu2019direct}.
    \item \textit{Efficiency Improvement}: Involves techniques such as network compression and sparse connectivity to reduce computational costs \citep{wu2019direct}.
    \item \textit{Temporal Dynamics Utilization}: Exploits sequential learning and integration with neuromorphic sensors to harness SNNs' unique temporal characteristics \citep{wu2019direct}.
\end{itemize}

Adaptive SNNs (AdSNNs) further advance SNNs \citep{zambrano2019sparse} by incorporating spike frequency adaptation, a phenomenon observed in biological neurons that dynamically modulates firing rates to encode input intensity effectively. The adaptation can be modeled through the dynamic threshold mechanism, where the threshold \(\vartheta(t)\) for spike generation adapts based on previous spiking activity. This adaptation can be expressed as:

\[
\vartheta_j(t) = \vartheta_0 + \sum_{t_j} m_f \vartheta_0 \gamma(t - t_j),
\]

where \(\vartheta_0\) is the resting threshold, \(m_f\) controls adaptation speed, and \(\gamma(t)\) is an adaptation kernel.

This adaptive mechanism allows for spike-based coding precision to be adjusted according to the input's dynamic range. By modulating firing rates, AdSNNs achieve efficient encoding while remaining energy-conscious, paralleling biological neural behavior. The flexibility in adjusting neural coding precision offers advantages in tasks that require varying levels of attentional focus, making AdSNNs suitable for temporally continuous, asynchronous applications.

\subsection{Distinctive Features of Our Approach}

While each of the methods discussed above enhances neural network adaptability and efficiency through unique mechanisms, our approach introduces a novel adaptive framework by dynamically adjusting neuron weights based on input signals. Unlike the \textit{adaptive neural networks for efficient inference} that rely on selecting pre-trained sub-networks, our model directly modifies weights at the neuron level to achieve real-time adaptability. In contrast to \textit{dynamic weighting in neural networks}, which generally incorporates fixed mechanisms like attention weights or conditionally parameterized convolutions, our model treats weights as functions of the input signal itself, enabling a more fluid response to input variations.

Our use of \textit{Chebyshev polynomials} is distinct from previous methods, such as that of Troumbis et al. \citep{troumbis2020chebyshev}, as it serves as an example within a broader framework that can integrate various orthogonal functions, not just Chebyshev polynomials. This design allows the network to be configured with any decomposition function, thereby expanding its flexibility across diverse applications.

Compared to \textit{spiking neural networks} (SNNs), which rely on binary spikes and temporal dynamics to mimic biological neurons, our approach achieves bio-inspired adaptability without the need for spike-based communication. Our model continuously adjusts weights as opposed to relying on discrete events, providing a smoother and potentially more computationally efficient pathway for tasks that do not necessitate spike-based signaling.

\subsection{Inspiration from Kolmogorov–Arnold Networks}
Multi-Layer Perceptrons (MLPs) \citep{rumelhart1986learning,rosenblatt1958perceptron,mcculloch1943logical}, commonly referred to as fully-connected feedforward neural networks, serve as essential components in modern deep learning architectures. MLPs consist of multiple layers of neurons, where each neuron applies a predefined activation function to the weighted sum of its inputs. This architecture enables MLPs to approximate a wide variety of nonlinear functions, a capability supported by the universal approximation theorem. Consequently, MLPs are widely employed in diverse deep learning tasks, including classification, regression, and feature extraction. However, MLPs are not without limitations, such as challenges in interpreting their learned representations and constraints in scaling the network.

Kolmogorov–Arnold Networks (KANs) \citep{liu2024kan} offer a novel alternative to conventional MLPs by utilizing the Kolmogorov-Arnold representation theorem. In contrast to MLPs, which use fixed activation functions for neurons, KANs introduce learnable activation functions on the edges, replacing linear weights with univariate functions modeled as splines.

The concept of adaptive neurons using Chebyshev polynomials was inspired by the Kolmogorov–Arnold Network (KAN) architecture, particularly its innovative approach of replacing fixed neuron activation functions with learnable functions on the edges. The key difference, however, lies in the representation of these adaptive weights. While KANs employ univariate spline functions to parameterize the weights, the proposed adaptive neurons use Chebyshev polynomials to model dynamic, input-dependent weights. This allows for a richer, more flexible representation of weight dynamics, enabling the neurons to adjust their output based on the input signal, thus offering a bio-inspired approach that more closely mirrors the behavior of biological neurons.

\section{Proposed Architecture}

In this section, we detail the architecture of the proposed adaptive neuron model using Chebyshev polynomials.

\subsection{Simple Neuron Model}

In a traditional neuron within a neural network, the process of generating the output relies on combining the inputs and their associated weights through a linear combination. The neuron takes a set of input values \( x_1, x_2, \ldots, x_n \), where each input \( x_i \) is multiplied by a corresponding weight \( w_i \). The weight represents the strength of the connection between the input and the neuron, determining how much influence that particular input will have on the final output.

Mathematically, the neuron computes a weighted sum of its inputs, which can be represented as:

\begin{equation}
y = \sum_{i=1}^{n} w_i \cdot x_i
\end{equation}

This equation describes a linear combination of the inputs, where each input \( x_i \) contributes to the output proportionally to its corresponding weight \( w_i \). The output of the neuron is simply the result of this linear combination, which is often passed through an activation function to introduce non-linearity and allow the network to learn more complex patterns.




In this basic model, the behavior of the neuron is limited to a linear combination of the inputs, meaning the output is only capable of representing linear functions of the inputs. To capture more complex relationships and patterns in the data, this output is typically passed through a non-linear activation function, such as the sigmoid or ReLU, which transforms the linear output into a non-linear space, allowing the neural network to model more sophisticated patterns.

This form of linear combination is the foundational operation in artificial neural networks, forming the basis for deeper architectures such as multi-layer perceptrons (MLPs), convolutional neural networks (CNNs), and other complex models.

\subsection{Decomposition of Weights Using Chebyshev Polynomials}

In traditional neural networks, the weights associated with each input to a neuron are fixed after training. However, biological neurons exhibit dynamic behavior where the strength of the connections (or synaptic weights) can adapt in response to stimuli. To better mimic this adaptive nature, we introduce a decomposition of the fixed weights into dynamic, input-dependent weights using Chebyshev polynomials.

Chebyshev polynomials are a family of orthogonal polynomials that provide a mathematically efficient way to approximate complex functions. By leveraging these polynomials, we enable the weights in our model to vary as a function of the input, making them more flexible and capable of capturing intricate, non-linear relationships between inputs and outputs. Specifically, instead of assigning a fixed weight \( w_i \) to an input \( x_i \), we represent \( w_i \) as a sum of Chebyshev polynomials, with the coefficients of these polynomials being learned during training.

For example, when using Chebyshev polynomials of order 3, the weight \( w_i \) corresponding to input \( x_i \) can be expressed as a sum of Chebyshev polynomials of increasing order. This is given by:

\begin{equation}
w_i (x_i) = \sum_{j=0}^{2} c_{i,j} T_j(x_i)
\label{eq:chebyshev_decomposition}
\end{equation}

Here, \( T_j(x_i) \) represents the Chebyshev polynomial of the first kind of degree \( j \), evaluated at input \( x_i \). The coefficients \( c_{i,j} \) are the learnable parameters that are optimized during training. By adjusting these coefficients, the network can adaptively alter the weight \( w_i \) based on the input \( x_i \).

\subsubsection{Chebyshev Polynomials of the First Kind}

Chebyshev polynomials of the first kind, \( T_j(x) \), are defined as:

\[
T_0(x) = 1, \quad T_1(x) = x, \quad T_2(x) = 2x^2 - 1, \quad T_3(x) = 4x^3 - 3x, \ldots
\]

These polynomials have useful properties, such as orthogonality, which allows for efficient representation and approximation of functions. When used to decompose the weights, these polynomials allow the model to dynamically adapt the contribution of each input in a non-linear and flexible manner.

\subsubsection{Generalization of the Decomposition}

We generalize the above formulation by extending it to all inputs in the network. For a set of inputs \( \{x_1, x_2, \ldots, x_n\} \), the adaptive weight for each input is expressed as:

\begin{equation}
w_i(x_i) = \sum_{j=0}^{k} c_{i,j} T_j(x_i)
\end{equation}

Thus, the neuron’s output, which is a weighted sum of all inputs, is given by:

\begin{equation}
Y = \sum_{i=1}^{n} x_i w_i(x_i) = \sum_{i=1}^{n} x_i \sum_{j=0}^{k} c_{i,j} T_j(x_i)
\end{equation}

This generalized formulation allows the weights for each input to vary dynamically, and the overall output reflects this flexibility.

\subsection{Chebyshev Polynomial Representation of Weights}

In this architecture, the input signals \( \{x_1, x_2, \ldots, x_n\} \) are transformed through Chebyshev polynomials \( T_j(x_i) \), and the adaptive weights are computed as a weighted sum of these polynomials. Each weight is expressed as:

\begin{equation}
w_i(x_i) = \sum_{j=0}^{k} c_{i,j} T_j(x_i)
\label{eq:chebyshev_weights_general}
\end{equation}

This representation allows each input weight to be dynamically adjusted based on the current input, thus enabling the network to adapt its responses.

\subsection{Swapping the Chebyshev Operation}

For computational efficiency, the Chebyshev operation can be swapped outside the neuron. This reorganization of the Chebyshev polynomial transformation allows for better modularity in the computation. The general form for the transformation can be written as:

\begin{equation}
T_j(x_i) \quad \text{precomputed and stored in advance,} \quad w_i(x_i) = \sum_{j=0}^{k} c_{i,j} T_j(x_i)
\end{equation}

\subsection{Encapsulating the Transformation}

To encapsulate the Chebyshev transformation into a reusable function, the model applies the Chebyshev transform to each input before it reaches the neuron. Mathematically, the transformation function can be represented as:

\begin{equation}
\text{Chebyshev Transform}(x_i) = \left[ T_0(x_i), T_1(x_i), \ldots, T_k(x_i) \right]
\end{equation}

\subsection{Fully Connected Network with Chebyshev Transformation}

Finally, we integrate the Chebyshev transformation into a fully connected network. The network processes inputs \( \{x_1, x_2, \ldots, x_n\} \) through the Chebyshev transformation, and the resulting adaptive weights are used for the final classification or regression tasks:

\begin{equation}
Y = \sum_{i=1}^{n} x_i \left( \sum_{j=0}^{k} c_{i,j} T_j(x_i) \right)
\end{equation}

\subsection{Adaptive Neuron Architecture with Chebyshev Polynomial Transformation}
\begin{figure}[htbp]
    \centering
    \includegraphics[width=0.8\textwidth]{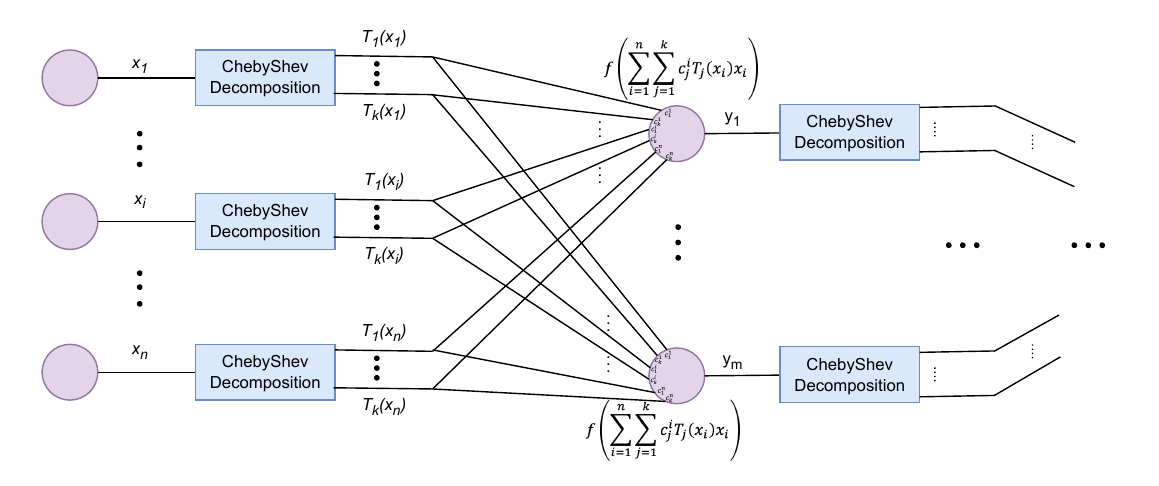}
    \caption{Experiment architecture with Chebyshev polynomial decomposition applied to each input \( x_i \). This transformation enables the model to adapt its responses dynamically by expanding each input feature through Chebyshev polynomial terms, improving the representation capability of the neuron model.}
    \label{fig:experiment_architecture}
\end{figure}

Figure \ref{fig:experiment_architecture} presents the architecture of the proposed adaptive neuron model, which incorporates Chebyshev polynomial transformations. In this design, each input feature \( x_i \) undergoes a Chebyshev decomposition, expanding it into a set of polynomial functions \( T_1(x_i), T_2(x_i), \ldots, T_k(x_i) \). This transformation allows for adaptive weighting of inputs, enhancing the model’s flexibility to capture complex patterns.

In the adaptive architecture, the Chebyshev decomposition is applied at the input layer, generating multiple transformed features for each input variable. This decomposition captures non-linear interactions, improving the model’s robustness. The polynomial terms are then combined through a set of learned coefficients, allowing the output \( y \) to adapt based on the input structure dynamically.

\subsection{Optimization of Adaptive Neural Networks}

Optimizing adaptive neural networks with decomposition-based weights introduces unique challenges, particularly in managing the number of parameters associated with each decomposition strategy. When using Gaussian decomposition, for example, two parameters—mean and standard deviation—are required for each decomposition component. Conversely, with Chebyshev polynomial decomposition, each term in the polynomial has only a single associated parameter, resulting in fewer parameters to optimize. This simplicity is one reason we favor Chebyshev polynomials in our adaptive neuron model.

The Gaussian decomposition for weight \( w_i(x_i) \) can be represented as:

\begin{equation}
w_i(x_i) = \sum_{j=0}^{k} c_{i,j} \exp \left(-\frac{(x_i - \mu_{i,j})^2}{2\sigma_{i,j}^2}\right)
\end{equation}

where \( c_{i,j} \) is the weight coefficient, \( \mu_{i,j} \) is the mean, and \( \sigma_{i,j} \) is the standard deviation. This requires the optimization of both \( \mu_{i,j} \) and \( \sigma_{i,j} \) for each term \( j \), increasing the parameter count and computational load.

For comparison, the Chebyshev decomposition has only one parameter per term:

\begin{equation}
w_i(x_i) = \sum_{j=0}^{k} c_{i,j} T_j(x_i)
\end{equation}

To illustrate the differences, Table \ref{tab:decomposition_comparison} provides a generalized comparison of the parameter requirements for different decomposition methods, assuming \( n \) features and polynomial order \( k \).

\begin{table}[htbp]
\centering
\begin{tabular}{p{4cm}p{4cm}p{5cm}}
\hline
\textbf{Decomposition Type} & \textbf{Parameters per Term} & \textbf{Total Parameters (n features, order k)} \\
\hline
Chebyshev Polynomial & \( c_{i,j} \) & \( n \times (k + 1) \) \\
Gaussian & \( c_{i,j}, \mu_{i,j}, \sigma_{i,j} \) & \( n \times 3(k + 1) \) \\
Fourier & \( a_{i,j}, b_{i,j} \) & \( n \times 2(k + 1) \) \\
Legendre Polynomial & \( c_{i,j} \) & \( n \times (k + 1) \) \\
\hline
\end{tabular}
\caption{Parameter requirements for different decomposition methods used in adaptive neural networks, with \( n \) as the number of features and \( k \) as the decomposition order.}
\label{tab:decomposition_comparison}
\end{table}

\subsubsection{Optimization of an Adaptive Neuron in Chebyshev Network}

In our adaptive Chebyshev neural network, each weight \( w_i \) is expressed as a function of Chebyshev coefficients, allowing dynamic adjustments based on the input. This decomposition provides the network with the flexibility to capture complex, non-linear relationships. Specifically, each weight \( w_i \) is represented as:

\[
w_i(x_i) = \sum_{j=0}^{k} c_{i,j} T_j(x_i)
\]

where \( c_{i,j} \) represents the Chebyshev coefficient for the \( j \)-th term, \( T_j(x_i) \) denotes the Chebyshev polynomial of order \( j \) evaluated at the input \( x_i \), and \( k \) is the highest polynomial order used. This decomposition makes each weight a flexible, input-dependent function, enhancing the model's adaptability.

To optimize these coefficients, we calculate the gradient of the loss \( L \) with respect to each Chebyshev coefficient \( c_{i,j} \), denoted as \( \Delta L_c \). This gradient calculation is structured in terms of two components: \(\Delta L_w\) and \(\Delta W_i\).

\begin{enumerate}
    \item Gradient of the Loss with respect to each Weight: The vector \(\Delta L_w\) represents the partial derivatives of the loss \( L \) with respect to each weight \( w_i \):

   \[
   \Delta L_w = 
   \begin{pmatrix}
   \frac{\delta L}{\delta w_{1}} \\
   \frac{\delta L}{\delta w_{2}} \\
   \vdots \\
   \frac{\delta L}{\delta w_{n}}
   \end{pmatrix}
   \]

   Here, \( \frac{\delta L}{\delta w_{i}} \) denotes how sensitive the loss function \( L \) is to changes in the weight \( w_i \). This term captures the impact of each weight on the overall loss.

   \item Gradient of each Weight with respect to Chebyshev Coefficients: For each weight \( w_i \), the vector \(\Delta W_i\) represents the partial derivatives of \( w_i \) with respect to each Chebyshev coefficient \( c_{i,j} \):

   \[
   \Delta W_i = 
   \begin{pmatrix}
   \frac{\delta w_i}{\delta c_{0,i}} \\
   \frac{\delta w_i}{\delta c_{1,i}} \\
   \vdots \\
   \frac{\delta w_i}{\delta c_{k,i}}
   \end{pmatrix}
   \]

   Here, \( \frac{\delta w_i}{\delta c_{j,i}} \) represents how sensitive the weight \( w_i \) is to changes in the \( j \)-th Chebyshev coefficient \( c_{i,j} \). This term quantifies how each coefficient contributes to the value of the weight \( w_i \).
\end{enumerate}

Each term in \(\Delta L_c\) is then computed as the product of the corresponding elements from \(\Delta L_w\) and \(\Delta W_i\), as shown below.

Thus, \(\Delta L_c\) can be expressed as:

\[
\Delta L_c = 
\begin{pmatrix}
\frac{\delta L}{\delta w_1} \cdot \Delta W_1 \\
\frac{\delta L}{\delta w_2} \cdot \Delta W_2 \\
\vdots \\
\frac{\delta L}{\delta w_n} \cdot \Delta W_n
\end{pmatrix}
\quad = \quad
\begin{pmatrix}
\frac{\delta L}{\delta w_1} \cdot \frac{\delta w_1}{\delta c_{0,1}} \\
\frac{\delta L}{\delta w_1} \cdot \frac{\delta w_1}{\delta c_{1,1}} \\
\vdots \\
\frac{\delta L}{\delta w_1} \cdot \frac{\delta w_1}{\delta c_{k,1}} \\
\frac{\delta L}{\delta w_2} \cdot \frac{\delta w_2}{\delta c_{0,2}} \\
\vdots \\
\frac{\delta L}{\delta w_2} \cdot \frac{\delta w_2}{\delta c_{k,2}} \\
\vdots \\
\frac{\delta L}{\delta w_n} \cdot \frac{\delta w_n}{\delta c_{0,n}} \\
\vdots \\
\frac{\delta L}{\delta w_n} \cdot \frac{\delta w_n}{\delta c_{k,n}} \\
\end{pmatrix}_{n(k+1) \times 1}
\]

This formulation allows us to compute \(\Delta L_c\) as a column vector of size \(n(k+1)\) by \(1\). The resulting gradient vector \(\Delta L_c\) can be utilized in standard backpropagation, enabling efficient training of the adaptive Chebyshev neural network.

By decomposing weights in terms of Chebyshev polynomials, this method offers a flexible and computationally efficient approach, reducing the number of parameters while enhancing the model's adaptability and capacity to represent complex, non-linear relationships.

\subsubsection{Preprocessing with Chebyshev Transformation}

\begin{figure}[htbp]
    \centering
    \includegraphics[width=0.8\textwidth]{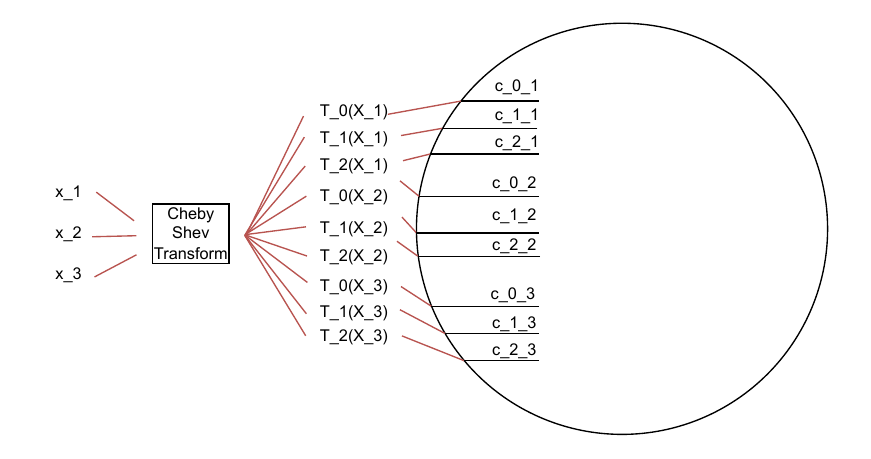}
    \caption{Illustration of a single operation unit where the Chebyshev transformation is applied as a preprocessing step to the input, preserving gradients and enabling efficient backpropagation.}
    \label{fig:operation_boxed}
\end{figure}

\begin{figure}[htbp]
    \centering
    \includegraphics[width=0.8\textwidth]{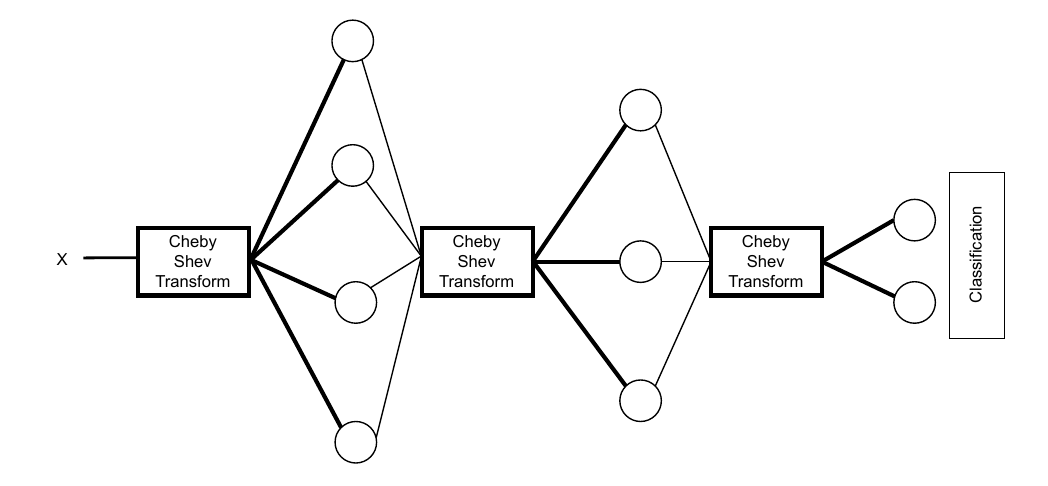}
    \caption{Full architecture demonstrating layer-wise preprocessing with Chebyshev transformation in an MLP model, facilitating standard backpropagation.}
    \label{fig:complete_architecture}
\end{figure}

In our model, the Chebyshev decomposition can be treated as a preprocessing step before each layer, which simplifies the backpropagation process. This approach enables us to leverage standard backpropagation techniques, as the Chebyshev transformation is applied to the input matrix before it is passed through the neuron layer. Figure \ref{fig:operation_boxed} shows how this transformation can be applied as a preprocessing step in a single operational unit, while Figure \ref{fig:complete_architecture} depicts the transformation at a layer-wise scale in an MLP architecture.

\subsection{Optimization Techniques for Adaptive Neural Networks}

Optimizing adaptive neural networks requires efficient techniques to handle the complexity introduced by decomposition-based adaptive weights. Various optimization methods have been proposed to enhance the convergence and stability of neural networks. \textbf{Gradient Descent} \citep{goodfellow2016deep} is a fundamental approach, adjusting weights by taking steps proportional to the negative gradient of the loss function. \textbf{Stochastic Gradient Descent (SGD)} \citep{bottou2010large} updates weights after each training example, speeding up convergence but increasing update variance. \textbf{Mini-Batch Gradient Descent} \citep{ruder2016overview} balances batch gradient descent and SGD by updating parameters after each mini-batch, improving convergence speed and reducing variance. Advanced techniques, such as \textbf{Momentum} \citep{qian1999momentum}, add a momentum term to reduce oscillations and accelerate convergence. \textbf{Nesterov Accelerated Gradient (NAG)} \citep{sutskever2013importance} builds upon Momentum by calculating gradients at a future point, enhancing convergence smoothness. \textbf{Adagrad} \citep{duchi2011adaptive} adapts learning rates individually for each parameter, though it suffers from decaying learning rates, which \textbf{AdaDelta} \citep{zeiler2012adadelta} addresses by accumulating only recent gradients. \textbf{Adam (Adaptive Moment Estimation)} \citep{kingma2014adam} combines aspects of both Momentum and Adagrad, maintaining exponentially decaying averages of past gradients and squared gradients, which improves convergence and robustness to hyperparameter tuning \citep{reddi2019convergence}. In our experiments, we employ the Adam optimizer, given its adaptability and efficiency, which are essential for training complex adaptive neural networks like our Chebyshev-based model.

\section{Data And Experiment}

\subsection{Datasets used}

For our experiments, we used 145 datasets from the Penn Machine Learning Benchmarks (PMLB) \citep{Olson2017PMLB}, a curated collection of benchmark datasets designed for evaluating and comparing supervised machine learning algorithms. These datasets cover a wide range of applications, including binary and multi-class classification, and they feature combinations of categorical, ordinal, and continuous features. For this study, we focused exclusively on the tabular classification datasets.

To provide an overview of the dataset characteristics, we generated visualizations that depict the distribution of various dataset properties, such as the number of rows, columns, and classes. These charts allow us to gain insights into the diversity and scale of the datasets used, helping us to understand the environments in which our proposed adaptive neuron model will be evaluated.

\begin{figure}[htbp]
  \centering
  \includegraphics[width=\textwidth]{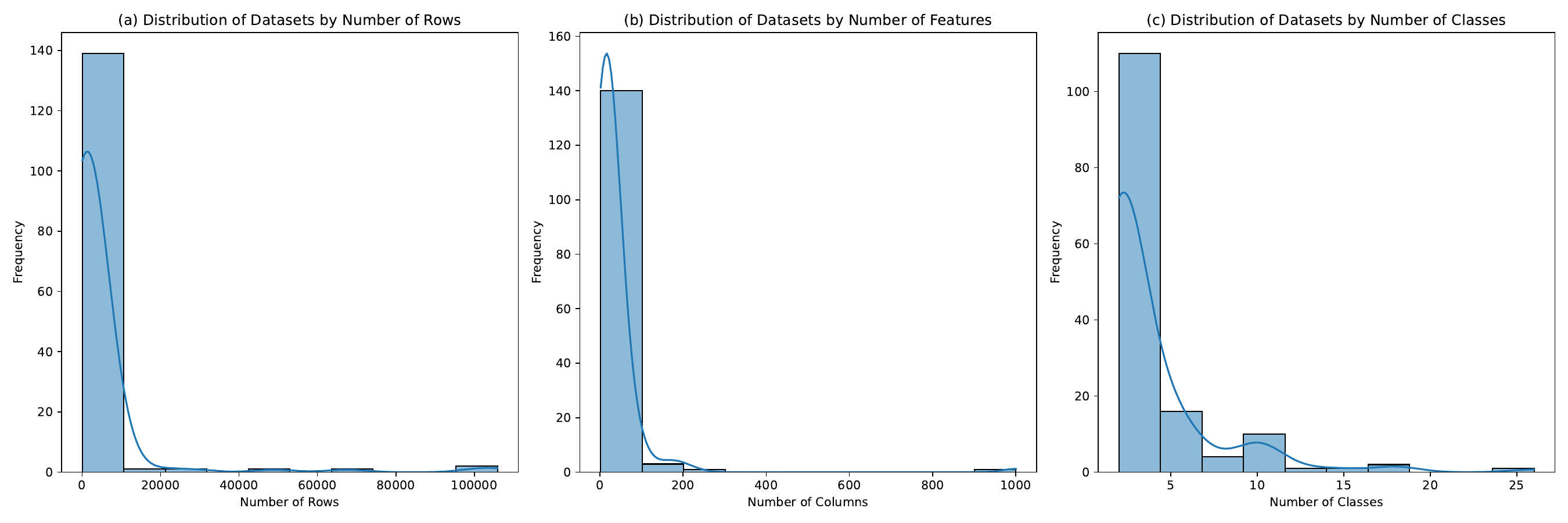}
  \caption{Overview of PMLB dataset distributions. (a) Distribution of datasets by the number of rows, highlighting the variability in dataset size, with most datasets containing a moderate number of rows suitable for robust model evaluation. (b) Distribution of datasets by the number of features, showcasing the range of dimensionalities that test the model's adaptability to various feature spaces. (c) Distribution of datasets by the number of classes, illustrating the diversity in classification tasks from binary to multi-class.}
  \label{fig:data_distribution}
\end{figure}

Figure \ref{fig:data_distribution}a illustrates the distribution of datasets by the number of rows, showing the frequency of datasets across different dataset sizes. This variation in dataset length provides a robust testing ground, especially for evaluating model scalability with larger datasets. Figure \ref{fig:data_distribution}b shows the distribution by the number of columns, representing the diversity in feature space dimensionality. This variety tests the model's capacity to handle datasets with different levels of complexity. Figure \ref{fig:data_distribution}c depicts the distribution by the number of classes, covering both binary and multi-class classification tasks. This diversity in target classes enables assessment of the model's performance across varying classification complexities. The datasets used in our experiments exhibit considerable diversity, making them well-suited for a comprehensive evaluation of the proposed adaptive neuron model. The range in dataset sizes is quite broad, with the largest dataset containing 105,908 rows and the smallest consisting of only 42 rows. Feature counts also vary significantly, spanning from 2 to 1000 features, while the number of target classes ranges from 2 to 26. This variety in dataset properties ensures that our model is tested across a wide spectrum of scenarios, from low-dimensional to high-dimensional feature spaces and from binary to complex multi-class classification tasks. Such diversity strengthens the evaluation, highlighting the model’s adaptability and robustness across diverse applications.

\subsection{Experiment Framework}

\begin{figure}[htbp]
  \centering
  \includegraphics[width=0.9\textwidth]{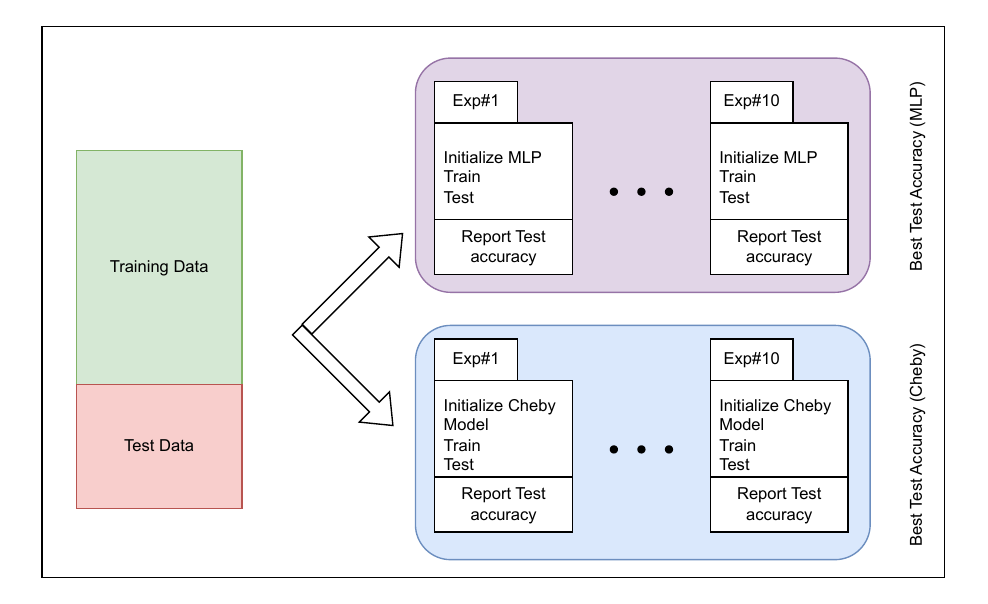}
  \caption{Workflow of the experiment showing the repeated training and evaluation steps for the MLP and Chebyshev models.}
  \label{fig:experiment_flow}
\end{figure}

The experiment evaluates the performance of a Multi-Layer Perceptron (MLP) and a Chebyshev-enhanced model across various datasets to compare their classification accuracies. We utilize a consistent model architecture for each dataset to ensure comparability.

For each dataset, we split the data into training and testing sets, followed by the training of both the MLP and Chebyshev models. Each model is independently trained 10 times on the same data split, and the best accuracy on the test set for each model is recorded for comparison. Figure \ref{fig:experiment_flow} illustrates the workflow of our experiment, showing the repeated training and evaluation steps for each model type.

\subsubsection{Model Architectures}

Both the MLP and Chebyshev-enhanced models share a consistent three-layer architecture for comparability. Each model has an initial layer with 4 neurons, followed by a second layer of 2 neurons, and an output layer where the number of neurons aligns with the number of classes in each dataset. The Chebyshev model applies a Chebyshev transformation to the input features before each layer, enhancing feature representation according to the Chebyshev order, but otherwise maintains the same structure as the MLP. The Chebyshev order, unless otherwise specified is set at 3.

\subsubsection{Training Parameters}

Both models are trained using consistent hyperparameters to ensure a fair comparison. A learning rate of 0.001 is applied, with the ReLU activation function guiding non-linear transformations within each layer. Optimization is handled by the Adam optimizer, chosen for its adaptive learning capabilities, and each model undergoes training over 500 epochs to allow sufficient convergence on the dataset patterns.

\subsection{Pruning to Leverage Higher-Order Chebyshev Terms}

We also explored the use of higher-order Chebyshev terms for model training, followed by parameter pruning to reduce model size while maintaining or even improving performance. We initially trained models with higher-order Chebyshev terms, up to degree 6, then applied pruning strategies to optimize model efficiency. Notably, the pruned models often achieved higher accuracy than the unpruned versions, as reported in the results section.
\subsubsection{Pruning Strategy 1: Parameter Thresholding}

Our initial pruning approach, denoted by the function \textbf{Prune}, involves setting a threshold to selectively remove parameters with values below a specific cut-off. In this straightforward method, each parameter’s value in the model is evaluated independently, without regard to its relationship within the Chebyshev decomposition structure. The pruning function is expressed as:

\[
W^{(1^*)} = \text{Prune} \left( W^{(1)} \mid W^{(2)}, \dots, W^{(m)} \right)
\]

where \( W^{(i)} \) represents the weights of layer $i$ before pruning and \( W^{(i*)} \) represents the weights of layer $i$ after pruning. Thus \( W^{(1^*)} \) represents the weights of layer $1$ after pruning, and \( m \) is the total number of layers in the network. This equation indicates that the pruning of \( W^{(1)} \) is performed independently of the other layers, focusing solely on the values within layer $1$ that fall below the threshold.

To further improve the pruning process, we define a function called \textbf{ForwardPrune} for subsequent layers. This function prunes weights in the current layer while taking into account the pruning state of previous layers. For example,

\[
W^{(2^*)} = \text{ForwardPrune} \left( W^{(2)} \mid W^{(1^*)}, W^{(3)}, \dots, W^{(m)} \right)
\]

where \( W^{(2^*)} \) represents the pruned weights of layer $2$, taking into account that layer $1$ has already been pruned. For any pruned layer (denoted with an asterisk *), weights set to zero remain frozen in the fine-tuning process, meaning they are not allowed to change, while non-zero weights continue to be fine-tuned. This approach ensures that once weights are set to zero, they stay inactive in subsequent training.

In general, the pruning process for any layer \( l \) can be expressed as:

\begin{equation}
W^{(l^*)} = \text{Prune} \left( W^{(l)} \mid W^{(1^*)}, W^{(2^*)}, \dots, W^{((l-1)^*)}, W^{(l+1)}, \dots, W^{(m)} \right)
\end{equation}

where \( m \) is the total number of layers in the network. This equation signifies that when pruning layer \( l \), the method considers all previously pruned layers to ensure a structured, layer-wise pruning approach. This method allows us to progressively prune the network, layer by layer, while taking into account the structure of previously pruned layers, thereby optimizing parameter reduction without disrupting the network's overall architecture.

\subsubsection{Pruning Strategy 2: Grouped Parameter Pruning for Chebyshev Decomposition}

In our experiments, we also introduced a novel pruning method tailored to the Chebyshev decomposition, which considers the structure of parameters within each Chebyshev polynomial expansion. For each feature \(i\), we compute a composite weight \(w_i\) that reflects the influence of all coefficients associated with that feature's Chebyshev terms. This is achieved by calculating the square root of the sum of squares for each coefficient \(c_{i,j}\) as follows:


\begin{equation}\label{eq:aggregate}
\| w_{i}(x) \|_2 \coloneqq \sqrt{\sum_{j=0}^{k} c_{i,j}^2}  
\end{equation}

where \(i\) is the feature index, \(j\) ranges from 0 to \(k\) (the highest Chebyshev polynomial degree), and \(c_{i,j}\) represents the coefficient for the \(j\)-th Chebyshev term in feature \(i\)'s expansion. Note that this calculation does not involve the input \(x\), focusing solely on the coefficients.

Following the calculation of \(w_i\) values, we apply the same threshold-based pruning function as in the first strategy. If any \(w_i\) in equation \ref{eq:aggregate} falls below the threshold, all coefficients \(c_{i,j}\) corresponding to that feature are pruned from the model. This approach ensures that low-impact features (as determined by their cumulative Chebyshev contributions) are removed in their entirety, enhancing the model's efficiency without compromising interpretability.

\subsubsection{Evaluation Methodology}

For each dataset, the following steps are conducted:
\begin{enumerate}
  \item The dataset is split into training and testing subsets.
  \item The MLP model is trained on the training subset 10 times, and the highest accuracy achieved on the test subset is recorded.
  \item The Chebyshev model undergoes the same training process on the same data split, with the highest accuracy on the test subset also recorded.
\end{enumerate}

Finally, the test accuracies for both the MLP and Chebyshev models are compared across all datasets, allowing us to assess the relative performance gains introduced by the Chebyshev enhancement.

\begin{figure}[htbp]
    \centering
    \includegraphics[width=0.8\textwidth]{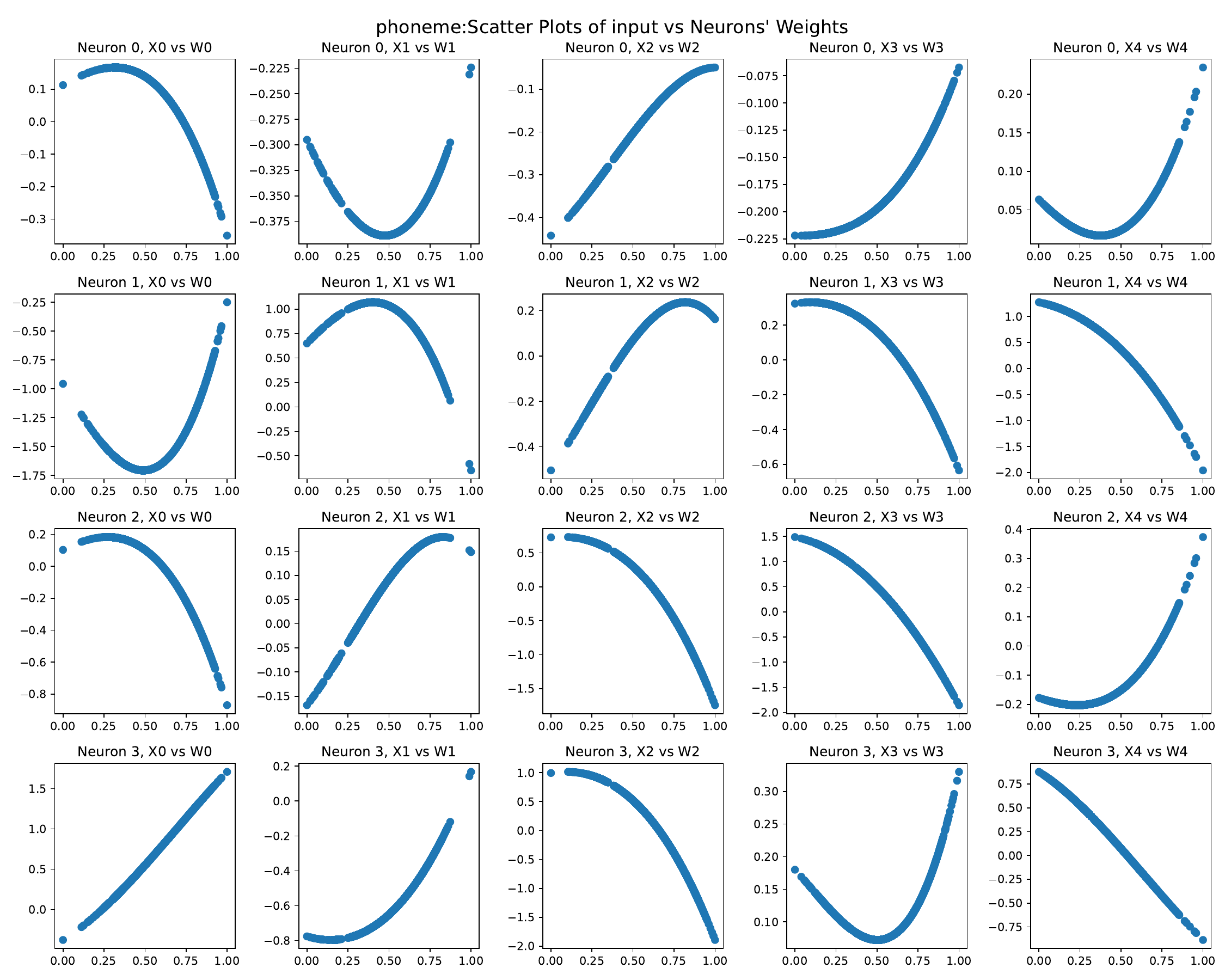}
    \caption{Scatter plots illustrating the relationship between inputs and adaptive weights for each neuron in the first layer, using the 'phoneme' dataset. The rows represent the neurons. At each row, the columns represent the weight distribution with respect to feature values. The smoothness in the weight transitions reflects the dynamic adaptability of weights based on varying input values.}
    \label{fig:phoneme_weights}
\end{figure}

\subsection{Analysis of Weight Distribution in Adaptive Neurons}

\begin{figure}[htbp]
    \centering
    \includegraphics[width=0.8\textwidth]{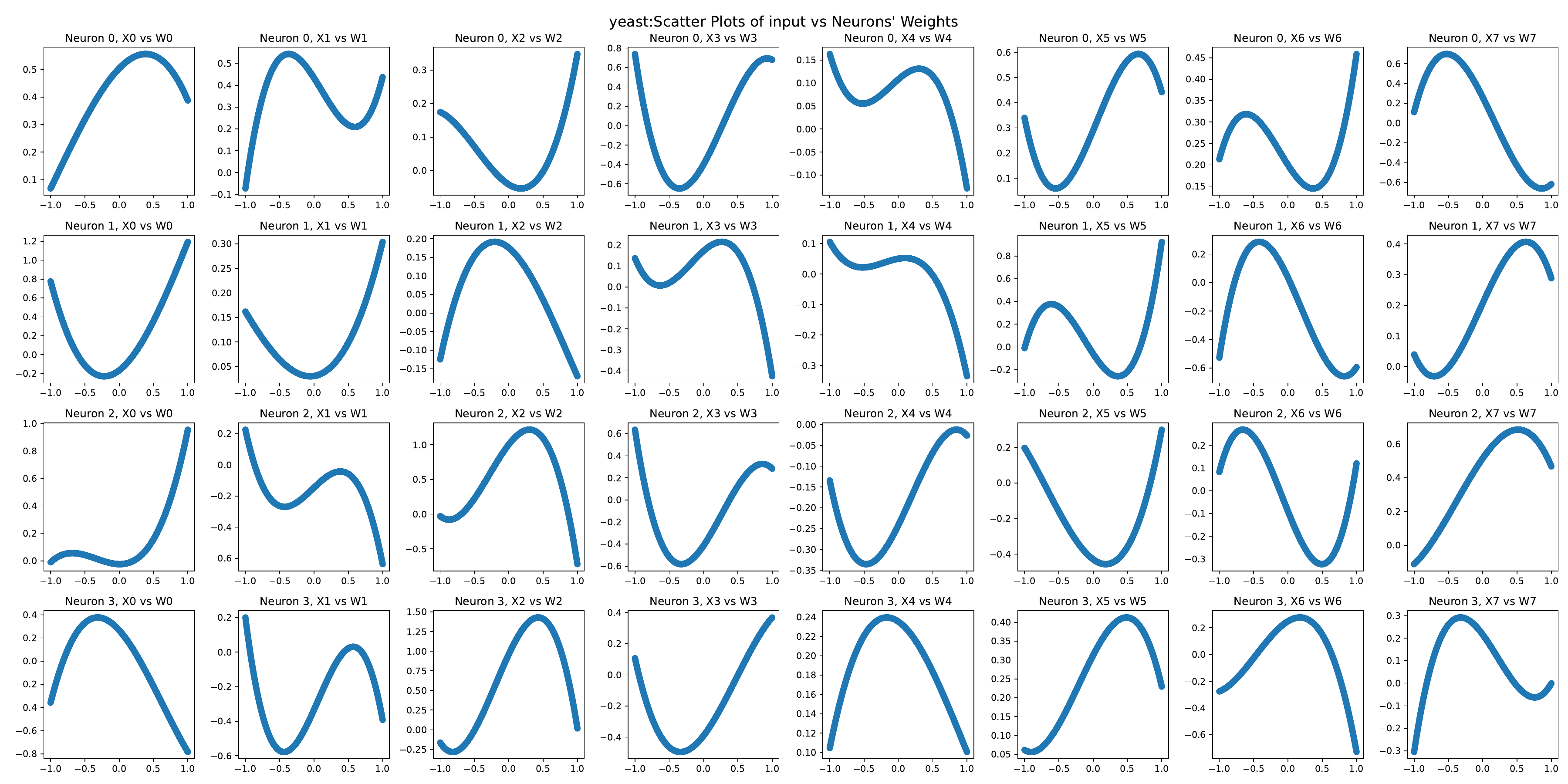}
    \caption{Scatter plots illustrating the relationship between inputs and adaptive weights for each neuron in the first layer, using the 'yeast' dataset. At each row, the columns represent the weight distribution with respect to feature values.. These plots further demonstrate the model’s capacity to adjust weights smoothly according to input changes.}
    \label{fig:yeast_weights}
\end{figure}

In this section, we examine how the weights in the adaptive neurons vary with changing input values, focusing on the first layer, which consists of 4 neurons. The weight distributions are analyzed and plotted for two representative datasets: 'phoneme' (Figure \ref{fig:phoneme_weights}) and 'yeast' (Figure \ref{fig:yeast_weights}). Each plot demonstrates the relationship between the input features and their corresponding adaptive weights, revealing the smooth transitions as inputs change. This behavior showcases the adaptability of the model in modulating weights based on input, which is instrumental in capturing non-linear patterns within the data.

This adaptive behavior enhances the network's ability to generalize across various datasets, enabling improved performance particularly in scenarios where input data exhibits complex, non-linear dependencies.

\begin{figure}[htbp]
  \centering
  \includegraphics[width=0.8\textwidth]{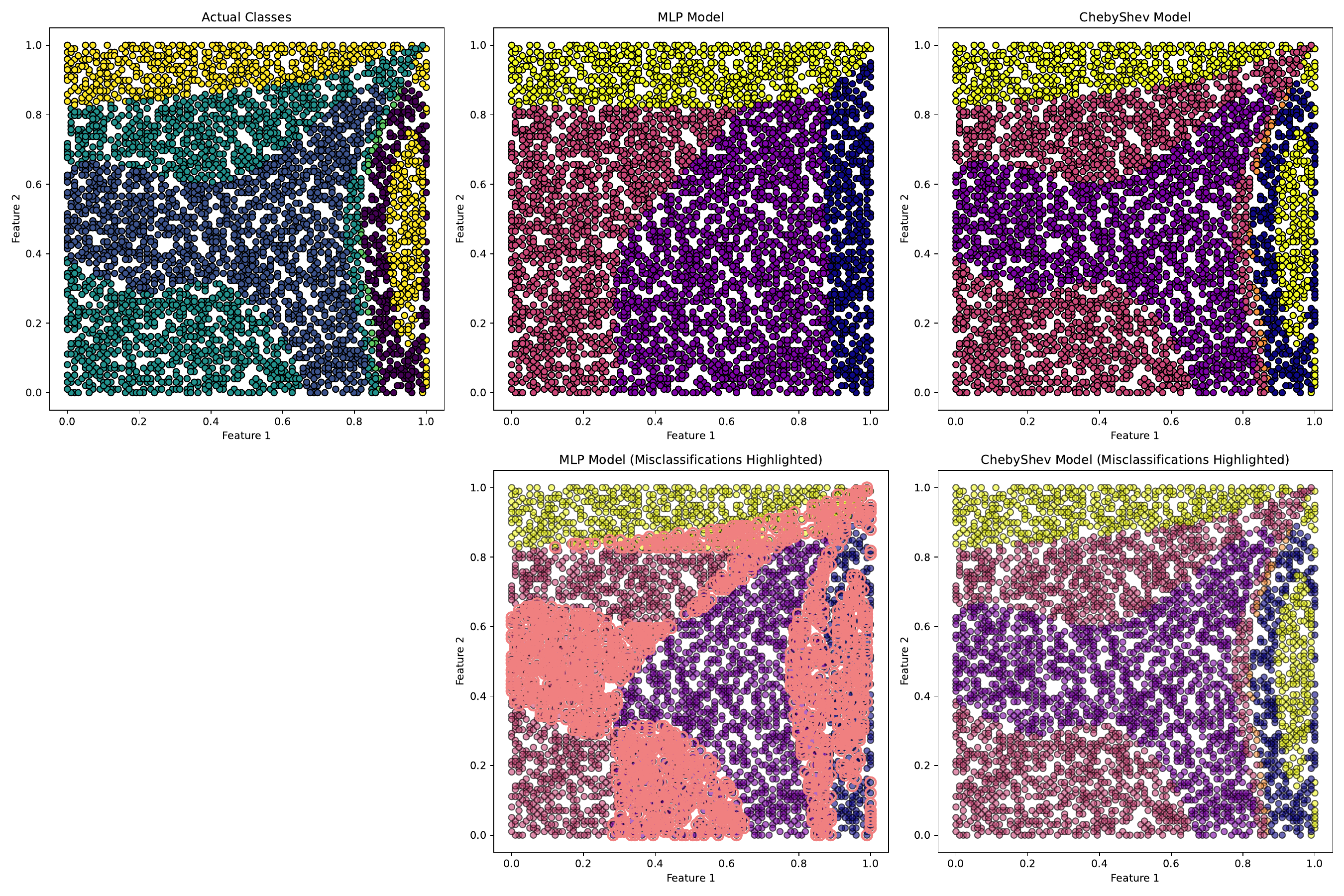}
  \caption{Comparison of decision boundaries on the "solar\_flare\_1" dataset with 5 classes. The first plot in the top row shows the actual class distribution. The second plot shows the MLP's decision boundaries, followed by Chebyshev's. The bottom row highlights misclassifications for each model, with pink areas indicating errors.}
  \label{fig:solar_flare_1_combined}
\end{figure}

\subsection{Analysis of Decision Boundaries}

To evaluate the ability of Chebyshev adaptive neural networks in learning decision boundaries, we analyzed their performance on challenging multiclass datasets. We trained both the Chebyshev model and the MLP on these datasets and observed the decision boundaries each model learned. Figure \ref{fig:solar_flare_1_combined} presents an example with the "solar\_flare\_1" dataset, which consists of 5 non-linearly distributed classes. The first plot in the first row illustrates the actual class distribution. The second and third plots in the first row show the decision boundaries learned by the MLP and Chebyshev models, respectively. Below each model's boundary plot, the corresponding misclassifications are highlighted in pink, providing insights into the models' abilities to correctly classify the data. The Chebyshev adaptive neural network demonstrates a clear advantage in capturing non-linear boundaries and reducing misclassifications.

\begin{figure}[htbp]
  \centering
  \includegraphics[width=0.8\textwidth]{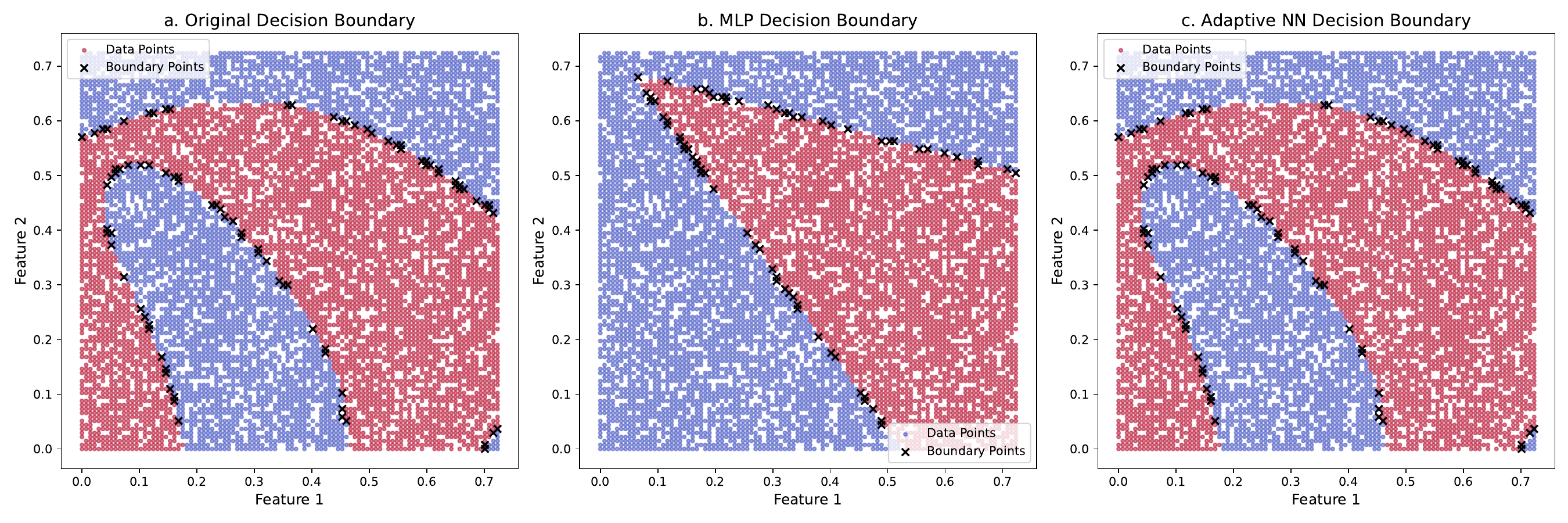}
  \caption{Decision boundary analysis on the "Hill Valley without noise" dataset. (a) Original scatter plot with decision boundary highlighted. (b) Decision boundary learned by the MLP, showing limited capacity to capture non-linearities. (c) Decision boundary learned by the Chebyshev model, effectively capturing non-linear boundaries.}
  \label{fig:decision_boundary_Hill_Valley_without_noise}
\end{figure}

The superiority of the Chebyshev model over the MLP is further illustrated in Figure \ref{fig:decision_boundary_Hill_Valley_without_noise}. Here, the decision boundary is highlighted in black. Figure \ref{fig:decision_boundary_Hill_Valley_without_noise}a shows the original data scatter plot with the true decision boundary, while Figure \ref{fig:decision_boundary_Hill_Valley_without_noise}b depicts the MLP’s decision boundary, which struggles to capture the non-linearities. In contrast, Figure \ref{fig:decision_boundary_Hill_Valley_without_noise}c shows the decision boundary learned by the Chebyshev model, which effectively captures the complex, non-linear boundaries.

These visualizations highlight the Chebyshev model's superior ability to learn complex, non-linear decision boundaries, underscoring its advantage over traditional MLPs in handling intricate data distributions.

Overall, the Chebyshev adaptive neural network demonstrates a strong capability in learning the decision boundaries of complex, non-linear datasets, outperforming the MLP in terms of boundary precision and reducing classification errors. Additional examples are provided in the supplementary section.

\section{Results and Discussion}

To assess the effectiveness of Chebyshev neural networks with adaptive weights compared to traditional Multi-Layer Perceptrons (MLP), we conducted a series of experiments across 145 datasets. 

\subsection{Comparative Accuracy of Chebyshev Adaptive Networks and MLPs}

\begin{figure}[htbp]
  \centering
  \includegraphics[width=0.55\textwidth]{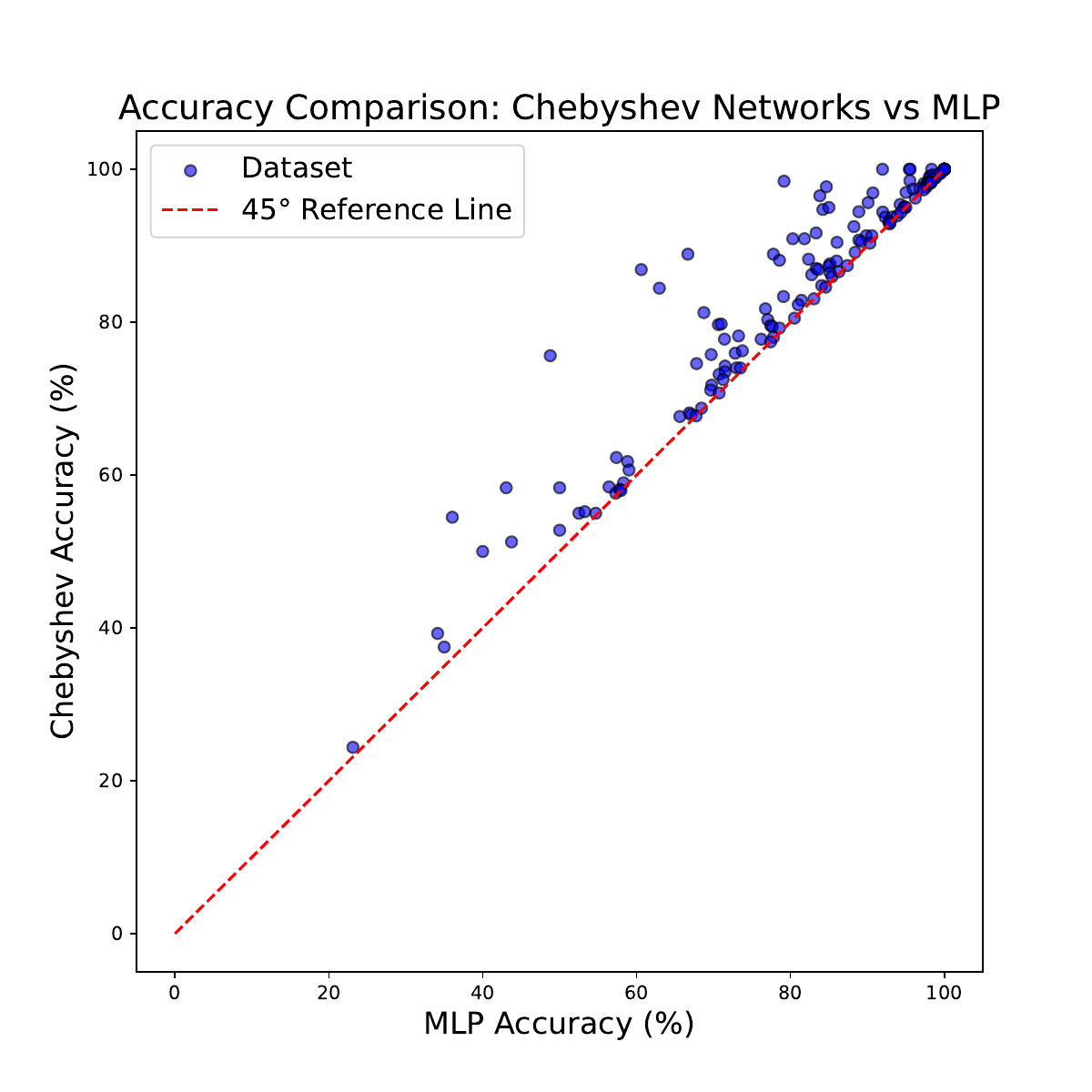}
  \caption{Scatter plot comparing the accuracy of Chebyshev Adaptive Networks and MLPs across 145 datasets. Each point represents a dataset, with points above the diagonal line indicating Chebyshev outperforming MLP.}
  \label{fig:scatter_plot}
\end{figure}

Figure \ref{fig:scatter_plot} presents a comprehensive comparison between Chebyshev Adaptive Networks and standard Multi-Layer Perceptrons (MLPs) across 145 datasets. Each point in the scatter plot corresponds to a dataset, where the x-axis represents MLP accuracy and the y-axis represents Chebyshev accuracy. The 45-degree diagonal line serves as a reference; points above the line indicate datasets where the Chebyshev model outperforms the MLP. The results demonstrate the superiority of the Chebyshev-based model in terms of generalization and predictive accuracy. In approximately 74\% of the datasets, the Chebyshev model achieves higher accuracy than the MLP, while in the remaining cases, it performs equivalently. This performance gain is particularly notable given that the Chebyshev model generalizes the MLP architecture and dynamically adapts its weights through polynomial transformations. 

These findings underscore the potential of adaptive neural networks, especially those leveraging Chebyshev polynomials, as robust alternatives to traditional MLPs—capable of capturing complex, non-linear relationships in data without increasing the parameter count.

\subsection{Task-Specific Insights: When Does the Chebyshev Model Perform Best?}

To understand which types of tasks or datasets benefit most from the Chebyshev Adaptive Network, we analyzed the top-performing datasets based on accuracy improvements over standard MLPs. Table~\ref{tab:all_data_desc} and Table~\ref{tab:cheby_mlp_accuracy_comparison} in appendix contain details about the datasets and the comprehensive results respectively. Table~\ref{tab:task_analysis} however, lists the datasets with the largest performance gains, along with their size, number of features, and number of classes as reported in the appendix.

\begin{table}[htbp]
\centering
\begin{tabular}{p{3.5cm}p{2cm}p{2cm}p{2cm}p{2cm}}
\hline
\textbf{Dataset} & \textbf{Gain (\%)} & \textbf{Size} & \textbf{\# Features} & \textbf{\# Classes} \\
\hline
auto & 26.83 & 121 & 25 & 5 \\
vowel & 26.26 & 594 & 13 & 11 \\
analcatdata fraud & 22.22 & 25 & 11 & 2 \\
soybean & 21.48 & 405 & 35 & 18 \\
tic tac toe & 19.27 & 574 & 9 & 2 \\
letter & 18.42 & 12000 & 16 & 26 \\
movement libras & 15.28 & 216 & 90 & 15 \\
ring & 13.04 & 4440 & 20 & 2 \\
car & 12.72 & 1036 & 6 & 4 \\
hayes roth & 12.50 & 96 & 4 & 3 \\
\hline
\end{tabular}
\caption{Top 10 datasets with the highest accuracy gain of Chebyshev model over MLP.}
\label{tab:task_analysis}
\end{table}

From this analysis, several trends emerge:

\begin{itemize}
    \item \textbf{High-dimensional datasets}: Datasets such as \textit{movement libras} (90 features) and \textit{soybean} (35 features) benefit significantly, suggesting the Chebyshev model’s ability to model complex feature interactions.
    \item \textbf{Multi-class classification}: Several of the top-performing datasets involve more than 4 classes (e.g., \textit{letter}, \textit{vowel}, \textit{soybean}), indicating that the Chebyshev model handles diverse class boundaries more effectively than MLPs.
    \item \textbf{Small to medium dataset sizes}: Many datasets have fewer than 1000 samples, which points to the Chebyshev model's generalization capabilities even under limited data.
\end{itemize}

These findings indicate that the Chebyshev Adaptive Network is particularly effective for tasks requiring expressive modeling of non-linear patterns, especially when dealing with high-dimensional inputs, multi-class settings, or limited data availability.

\subsection{Error Analysis: Understanding Performance Limitations}
While our proposed adaptive Chebyshev neural network outperformed traditional MLPs on 74\% of the datasets, it is important to understand why it did not outperform in the remaining 26\%. To this end, we performed an error analysis by investigating the datasets where our method failed to surpass the MLP baseline.

Upon inspection, several common characteristics were observed among the underperforming datasets:

\begin{itemize}
    \item \textbf{Very small dataset sizes:} Many datasets where Chebyshev underperformed contained fewer than 100 samples (e.g., \textit{analcatdata asbestos}, \textit{hepatitis}). In such cases, the added model complexity from the Chebyshev decomposition could lead to overfitting.
    \item \textbf{Low number of features:} A large subset of underperforming datasets had fewer than 5 input features, reducing the benefit of applying higher-order polynomial expansions.
    \item \textbf{Binary classification tasks with linearly separable features:} In cases where data was already well-separated using linear boundaries, the adaptive transformation introduced unnecessary complexity without significant benefit.
\end{itemize}

These findings suggest that the proposed model may be less effective on datasets that are small, low-dimensional, or linearly separable, where the expressive power of dynamic polynomial weighting is either unneeded or prone to overfitting. As part of future work, we plan to introduce a model selection mechanism to adaptively adjust the Chebyshev order \(k\) based on dataset complexity.

\subsection{Comparison with MLPs of Equal or Greater Parameter Count}\label{sub:comp_MLP_params}

To ensure a fair and meaningful comparison, we configured the baseline Multi-Layer Perceptrons (MLPs) with the same or a higher number of trainable parameters than the Chebyshev adaptive networks. Table~\ref{tab:top_chebyshev_gains} highlights the top 9 datasets where Chebyshev models demonstrated the largest improvements in F1-score.

Despite the MLPs having equal or greater capacity in terms of parameter count, the Chebyshev networks consistently achieved higher F1-scores in all the datasets. This illustrates that the performance improvements are not a result of increased model size, but stem from the adaptive and expressive capabilities introduced by the Chebyshev polynomial transformation.

The use of the F1-score—a balanced metric that accounts for both precision and recall—further emphasizes the robustness of our model across datasets with varying class distributions. The results indicate that even when controlling for model capacity, the Chebyshev adaptive network delivers superior generalization, highlighting its effectiveness as a strong alternative to traditional MLP architectures.

\begin{table}[htbp]
\centering
\begin{tabular}{p{3cm}p{3cm}p{4cm}p{3cm}}
\hline
\textbf{Dataset} & \textbf{Chebyshev F1-Score (\# parameters)} & \textbf{MLP F1-Score (\# parameters)} & \textbf{Improvement} \\
\hline
analcatdata fraud & 60.51 (8322) & 39.68 (8546) & 20.83 \\
hayes roth & 77.87 (7011) & 68.75 (7235) & 9.12 \\
auto & 74.8 (11109) & 66.0 (11333) & 8.8 \\
movement libras & 86.36 (23919) & 80.5 (24143) & 5.86 \\
car & 97.67 (7428) & 92.38 (7652) & 5.29 \\
tic tac toe & 95.83 (7938) & 92.14 (8162) & 3.69 \\
soybean & 89.42 (13458) & 87.09 (13682) & 2.33 \\
ring & 96.82 (10050) & 96.62 (10274) & 0.2 \\
vowel & 94.5 (9003) & 94.46 (9227)& 0.04 \\

\hline
\end{tabular}
\caption{F1-score comparison on 9 representative datasets where the Chebyshev Adaptive Network significantly outperforms or matches MLPs, despite the latter having equal or greater parameter counts. This demonstrates the effectiveness of the Chebyshev architecture beyond model size.}
\label{tab:top_chebyshev_gains}
\end{table}

These findings underscore the potential of the Chebyshev neural network with adaptive weights, demonstrating its enhanced accuracy across a diverse set of datasets, especially in cases with non-linear patterns that may not be effectively captured by traditional MLP architectures.

\subsection{Effect of value of k: Overfitting}

\begin{figure}[htbp]
    \centering
    \includegraphics[width=0.7\textwidth]{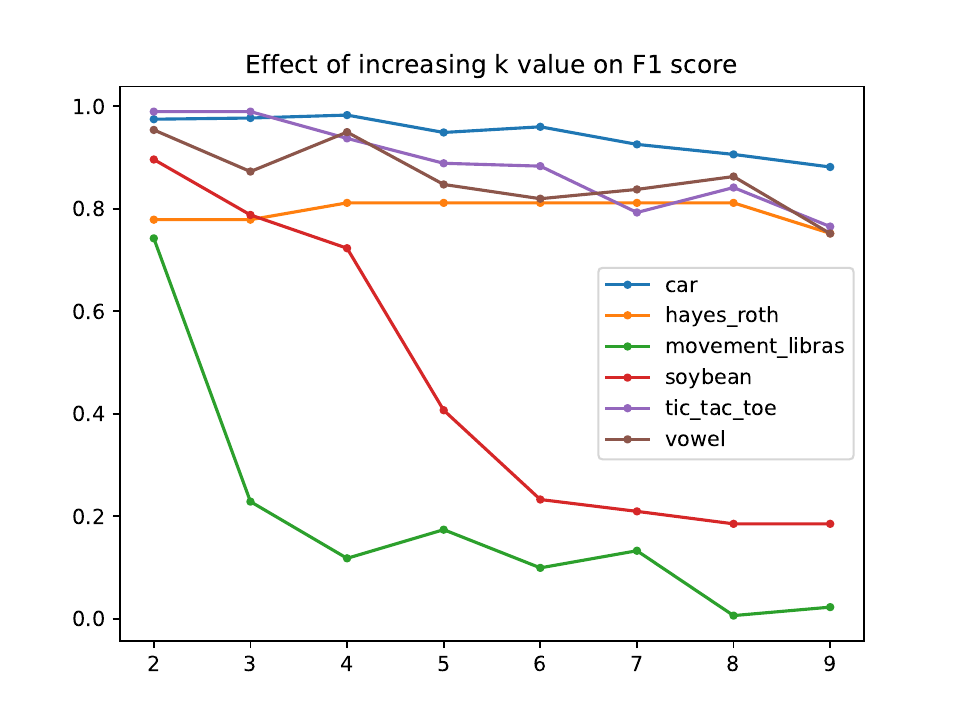}
    \caption{Effect of increasing Chebyshev order $k$ on the test F1 score across five datasets. While moderate values of $k$ improve performance, excessive values lead to overfitting and degraded generalization.}
    \label{fig:inc_k_f1}
\end{figure}

A critical hyperparameter in our adaptive Chebyshev neural network is the polynomial order $k$, which determines the number of Chebyshev basis functions used to model the input-dependent weights. While higher-order polynomials offer increased flexibility and modeling capacity, they can also introduce risks such as overfitting, instability during training, and degraded generalization—particularly on small or noisy datasets.

To explore the influence of $k$, we conducted experiments across several datasets and plotted the change in test F1 score with increasing values of $k$. As shown in Figure~\ref{fig:inc_k_f1}, we observe a consistent trend: performance initially improves with increasing $k$, reaching a peak, and subsequently declines as overfitting sets in. This is especially apparent in datasets such as \textit{hayes\_roth} and \textit{car}, where a high $k$ leads to sharp drops in F1 score.

This behavior confirms that while the adaptive Chebyshev network generalizes the MLP (which corresponds to $k=0$), an improperly chosen $k$ can lead to overparameterization and poor generalization. The increase in learnable parameters, introduced by higher-order Chebyshev terms, allows the network to fit the training data too closely—especially in low-data regimes—resulting in diminished test performance.

\subsection{Computational Efficiency Analysis}

We measured the training time per batch and inference time per batch for four neural architectures—
\begin{itemize}
    \item Multi-Layer Perceptron (MLP),
    \item Attention-based \citep{vaswani2017attention} FCNN (AttentionFCNN),
    \item Hypernetwork-based \citep{ha2016hypernetworks} FCNN (HyperFCNN), and
    \item Chebyshev-enhanced FCNNs (ChebyFCNN) with polynomial orders k [2, 4, 6 and 8]
\end{itemize}

across synthetic datasets of increasing feature dimensionality (4, 6, 9, 11, 16, 20, 35, and 90). Each result represents the average runtime for a single batch over 30 repetitions, after warm-up iterations, measured separately for training (forward + backward + update) and inference (forward only).
\subsubsection{Training Time}

\begin{figure}[htbp]
    \centering
    \includegraphics[width=0.9\textwidth]{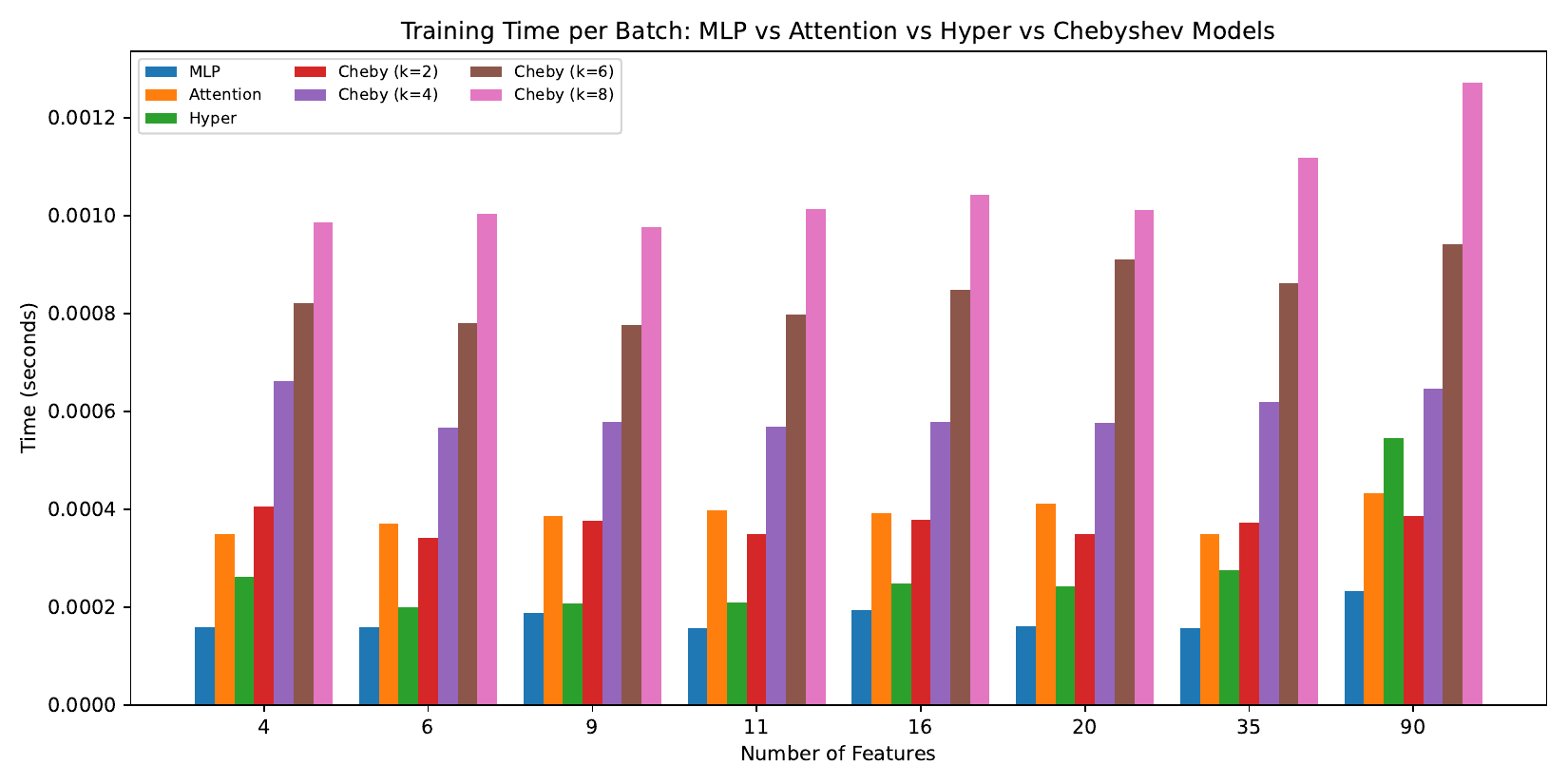}
    \caption{Training time per batch for MLP, AttentionFCNN, HyperFCNN, and ChebyFCNN with polynomial orders k = 2, 4, 6, 8 across datasets of varying feature dimensionality (4–90). MLPs achieve consistently minimal training costs, while Chebyshev-based models scale in runtime with both input dimensionality and polynomial order.}
    \label{fig:training_time_bar}
\end{figure}

MLPs were consistently the most efficient, requiring only ~0.00016–0.00023 seconds per batch across all feature sizes. As seen in Figure \ref{fig:training_time_bar} at 90 features, AttentionFCNNs trained about 1.9× slower than MLPs, HyperFCNNs about 2.3×, and Chebyshev models scaled more steeply with polynomial order: 1.7× (k=2), 2.8× (k=4), 4.0× (k=6), and 5.4× (k=8) compared to MLPs.

\subsubsection{Inference Time}

\begin{figure}[htbp]
    \centering
    \includegraphics[width=0.9\textwidth]{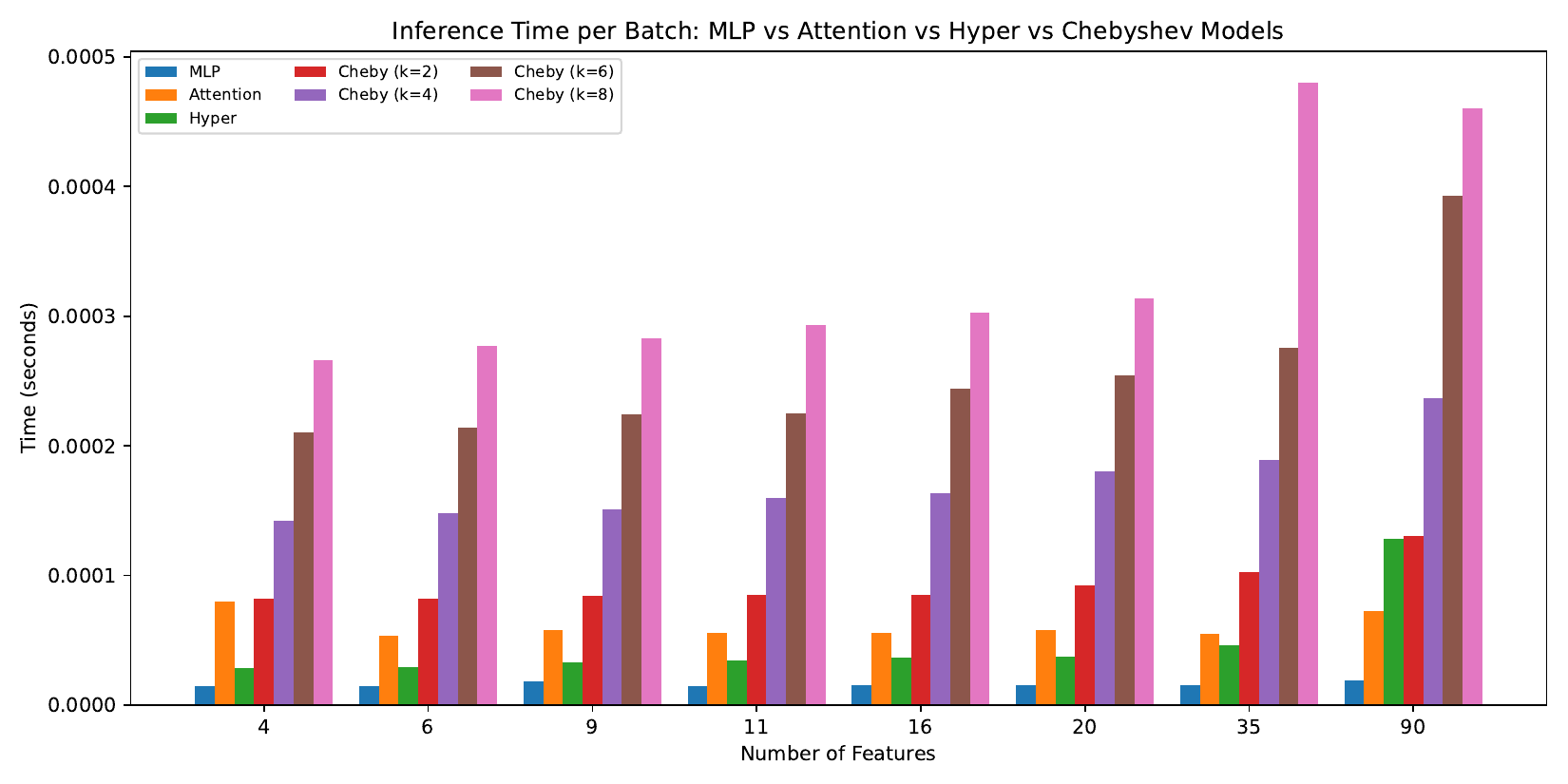}
    \caption{Inference time per batch for the same models and settings. MLPs are nearly instantaneous, whereas AttentionFCNN and HyperFCNN add modest overhead. Chebyshev-based models show significantly higher inference times, especially for larger orders, reflecting the computational expense of polynomial feature expansions.}
    \label{fig:inference_time_bar}
\end{figure}

A similar trend was observed for inference as seen in Figure \ref{fig:inference_time_bar}. MLPs completed forward passes in ~0.000015–0.000019 seconds. At 90 features, AttentionFCNNs were 3.8× slower, HyperFCNNs 6.7× slower, while Chebyshev variants ranged from 6.8× (k=2) to 23.9× (k=8) slower than MLPs.

These results highlight clear efficiency trade-offs.
\begin{itemize}
    \item MLPs provide unmatched speed but no dynamic input-dependent computation.
    \item Attention and Hypernetworks introduce modest overheads while enriching adaptability.
    \item Chebyshev-based networks incur substantially higher computational cost, particularly at larger orders, due to repeated polynomial expansions and feature interactions.
\end{itemize}

Nevertheless, these figures highlight the trade-off between the computational efficiency of MLPs and the expressive adaptability of Chebyshev-based neurons, justifying their increased runtime when richer input-conditioned representations are desired.

\subsection{Performance after Pruning}
The results in Table \ref{tab:pruned_chebyshev_gains} illustrate the performance of the pruned Chebyshev adaptive neural network across  datasets, with datasets ordered by decreasing compression levels. We show the 7 datasets where maximum compression could be achieved. The rest of the table can be found in the appendinx section. Remarkably, the pruning process allowed for up to ~90\% compression in model size, accompanied by an increase in accuracy compared to the original MLP performance. In cases where the unpruned Chebyshev model accuracy matched that of the MLP, pruning enabled a boost in accuracy, demonstrating the effectiveness of this approach in both reducing model complexity and enhancing performance.

\begin{table}[htbp]
\centering
\begin{tabular}{p{3cm}p{2cm}p{3cm}p{3cm}}
\hline
\textbf{Dataset} & \textbf{MLP Accuracy} & \textbf{Chebyshev Adaptive Model Accuracy (Pruned)} & \textbf{Compression} \\
\hline
wdbc & 98.246 & 99.123 & 89.2 \\
\hline
credit a & 84.057 & 84.783 & 89.1 \\
\hline
tokyo1 & 93.229 & 93.75 & 86.3 \\
\hline
breast & 100 & 100 & 85.8 \\
\hline
pima & 78.571 & 79.221 & 82.9 \\
\hline
cmc & 57.288 & 57.627 & 80.3 \\
\hline
cleveland nominal & 59.016 & 60.656 & 79.7 \\
\hline
\end{tabular}
\caption{Performance of pruned Chebyshev adaptive neural networks compared to standard MLPs across various datasets. The table highlights improvements in accuracy achieved by pruning, with the datasets arranged in decreasing order of model compression. For datasets where the unpruned Chebyshev model matched MLP performance, pruning not only increased accuracy but also resulted in substantial model compression, achieving up to ~90\% reduction in model size while enhancing accuracy. The table columns display the dataset names, MLP accuracy, pruned Chebyshev model accuracy, and the compression percentage.}
\label{tab:pruned_chebyshev_gains}
\end{table}

\subsection{Comparison with other architectures like self-attention and hypernetworks}

To further contextualize the effectiveness of Chebyshev Adaptive Networks, we extended our comparison to other architectures that also rely on input-dependent weight computations, namely self-attention \citep{vaswani2017attention} and hypernetworks \citep{ha2016hypernetworks}. For this purpose, we designed two baseline models: Attention Network (AttentionFCNN), which uses an embedding layer followed by a multi-head self-attention mechanism and a fully connected classifier, and Hyper Network (HyperFCNN), where a hypernetwork dynamically generates the classifier’s weight matrix and bias for each input. Both of these models represent alternative approaches to capturing richer, data-dependent feature interactions that go beyond static parameterizations. For a fair comparison, we adjusted hidden layer sizes to ensure parameter counts comparable to those of the Chebyshev models. We then evaluated all models on a set of representative PMLB datasets, reporting the best test F1-scores alongside their parameter counts.

\begin{table}[htbp]
\centering
\begin{tabular}{p{3cm}p{3cm}p{3cm}p{3cm}}
\hline
\textbf{Dataset} & \textbf{Chebyshev (\# params)} & \textbf{AttentionFCNN (\# params)} & \textbf{HyperFCNN (\# params)} \\
\hline
analcatdata fraud & 60.51 (8322)   & 58.46 (8482)  & \textbf{75.00} (8570) \\
hayes roth        & 77.87 (7011)   & 75.24 (7431)   & \textbf{80.02} (7171) \\
movement libras   & \textbf{86.36} (23919)  & 84.14 (24161)  & 80.09 (24545) \\
tic tac toe       & 95.83 (7938)   & 90.76 (8186)   & \textbf{97.72} (8018) \\
soybean           & 89.42 (13458)  & 91.85 (14050)  & \textbf{92.47} (13870) \\
ring              & \textbf{96.82} (10050)  & 81.20 (10520)  & 96.69 (10336) \\
vowel             & 94.50 (9003)   & 97.38 (9299)   & \textbf{97.63} (9281) \\
\hline
\end{tabular}
\caption{F1-score comparison across seven representative datasets for Chebyshev Adaptive Networks, AttentionFCNN, and HyperFCNN. The table reports both the F1-score and the number of parameters, showing that Chebyshev-based models remain competitive with other input-dependent architectures across different dataset types.}
\label{tab:cheby_attention_hyper_comparison}
\end{table}

The results, shown in Table \ref{tab:cheby_attention_hyper_comparison}, highlight that Chebyshev networks remain competitive against attention and hypernetwork baselines across diverse datasets. In some cases, Chebyshev models outperform their counterparts, while in others the dynamic attention or hypernetwork mechanisms achieve stronger results. Together with the earlier analysis in \ref{sub:comp_MLP_params}, where Chebyshev models consistently surpassed traditional fully connected networks, these findings demonstrate that Chebyshev-based architectures form a robust competitor to other advanced input-dependent methods.

\section{Future Directions}

Future work could explore the application of adaptive weights in out-of-distribution scenarios, focusing on their potential to enhance model resilience and improve generalization beyond standard datasets. Additionally, alternative decomposition methods, such as Fourier, Legendre, and Hermite polynomials, could be investigated to assess how different decomposition techniques impact model performance across various dataset characteristics. By examining these alternatives, research could identify which decomposition types are best suited for specific dataset attributes, including class count, class imbalance, and data distribution properties, thereby enabling more targeted applications of adaptive models.

For functions involving multiple variables, \( f(x_1, x_2, \ldots, x_d) \), Chebyshev decomposition can be extended to higher dimensions through a multivariate Chebyshev series expansion. This approach approximates \( f(x_1, x_2, \ldots, x_d) \) using products of univariate Chebyshev polynomials for each variable, providing a powerful tool for multivariate function approximation. 

The following is a step-by-step outline for decomposing a multivariate function using Chebyshev polynomials, which forms the foundation for future exploration in high-dimensional adaptive networks.

\subsection{Steps for Multivariate Chebyshev Decomposition}

To extend Chebyshev decomposition to multivariate functions, we start by defining the multivariate Chebyshev polynomial basis. For \( d \) dimensions, the basis functions are products of univariate Chebyshev polynomials for each variable, so for a given degree \( m_i \) in each dimension \( x_i \), the basis function is \( T_{m_1, m_2, \ldots, m_d}(x_1, x_2, \ldots, x_d) = T_{m_1}(x_1) T_{m_2}(x_2) \cdots T_{m_d}(x_d) \), where \( T_{m_i}(x_i) \) represents the Chebyshev polynomial of degree \( m_i \) in the \( i \)-th variable, defined on \([-1, 1]\). If the domain of \( f(x_1, x_2, \ldots, x_d) \) lies outside \([-1, 1]^d\), we map each variable \( x_i \) from \([a_i, b_i]\) to \([-1, 1]\) using \( x_i' = \frac{2x_i - (a_i + b_i)}{b_i - a_i} \), scaling and shifting \( x_i \) so that \( f(x_1, x_2, \ldots, x_d) \) is transformed to \( f(x_1', x_2', \ldots, x_d') \) on \([-1, 1]^d\). We then approximate \( f(x_1, x_2, \ldots, x_d) \) by a finite multivariate Chebyshev series: 

\[
f(x_1, x_2, \ldots, x_d) \approx \sum_{m_1=0}^{M_1} \sum_{m_2=0}^{M_2} \cdots \sum_{m_d=0}^{M_d} c_{m_1, m_2, \ldots, m_d} T_{m_1}(x_1) T_{m_2}(x_2) \cdots T_{m_d}(x_d),
\]

where \( c_{m_1, m_2, \ldots, m_d} \) are the Chebyshev coefficients. These coefficients are calculated as 

\[
c_{m_1, m_2, \ldots, m_d} = \frac{2^d}{\pi^d} \int_{-1}^{1} \cdots \int_{-1}^{1} \frac{f(x_1, x_2, \ldots, x_d) T_{m_1}(x_1) \cdots T_{m_d}(x_d)}{\sqrt{1 - x_1^2} \cdots \sqrt{1 - x_d^2}} \, dx_1 \cdots dx_d,
\]

but due to computational difficulty, these integrals are usually approximated using Chebyshev nodes and discrete transforms. For each variable \( x_i \), we choose \( M_i + 1 \) Chebyshev nodes as \( x_i^{(k)} = \cos\left(\frac{(2k + 1) \pi}{2(M_i + 1)}\right) \), for \( k = 0, 1, \dots, M_i \), and evaluate \( f \) at each point on the resulting \( (M_1 + 1) \times (M_2 + 1) \times \cdots \times (M_d + 1) \) grid, yielding \( f_{k_1, k_2, \ldots, k_d} = f\left(x_1^{(k_1)}, x_2^{(k_2)}, \ldots, x_d^{(k_d)}\right) \). To find the coefficients \( c_{m_1, m_2, \ldots, m_d} \), we then apply a multidimensional Discrete Cosine Transform (DCT) to these sampled values. Finally, the multivariate Chebyshev polynomial approximation for \( f(x_1, x_2, \ldots, x_d) \) is constructed as 

\[
f(x_1, x_2, \ldots, x_d) \approx \sum_{m_1=0}^{M_1} \sum_{m_2=0}^{M_2} \cdots \sum_{m_d=0}^{M_d} c_{m_1, m_2, \ldots, m_d} T_{m_1}(x_1) T_{m_2}(x_2) \cdots T_{m_d}(x_d).
\]

\subsection{Example: 3D Chebyshev Decomposition with Pairwise Combinations}

For a function of three variables, \( f(x, y, z) \), a pairwise Chebyshev decomposition can help in capturing interactions between pairs of variables. This involves decomposing \( f(x, y, z) \) as a sum of functions of pairs of variables:

\[
f(x, y, z) \approx \sum_{m=0}^{M} \sum_{n=0}^{N} c_{m,n}^{(x,y)} T_m(x) T_n(y) + \sum_{m=0}^{M} \sum_{p=0}^{P} c_{m,p}^{(x,z)} T_m(x) T_p(z) + \sum_{n=0}^{N} \sum_{p=0}^{P} c_{n,p}^{(y,z)} T_n(y) T_p(z),
\]

where \( c_{m,n}^{(x,y)}, c_{m,p}^{(x,z)}, \) and \( c_{n,p}^{(y,z)} \) are the Chebyshev coefficients for each pairwise combination.

\subsection{Efficient Multivariate Function Approximation Using Pairwise Chebyshev Decomposition}

We aim to classify a dataset with features \( x_1, x_2, \ldots, x_d \), with the objective of predicting \( y \) based on the function \( y = f(x_1, x_2, \ldots, x_d) \). The question is how to estimate the function \( f(x_1, x_2, \ldots, x_d) \). Using function decomposition, we can represent \( f(x_1, \ldots, x_d) \) with orthogonal functions \( T_{m_1, m_2, \ldots, m_d}(x_1, \ldots, x_d) \) as follows:

\[
f(x_1, x_2, \ldots, x_d) = \sum_{m_1=0}^{\infty} \sum_{m_2=0}^{\infty} \cdots \sum_{m_d=0}^{\infty} c_{m_1, m_2, \ldots, m_d} T_{m_1, m_2, \ldots, m_d}(x_1, x_2, \ldots, x_d),
\]

or

\[
f(x_1, x_2, \ldots, x_d) \approx \sum_{m_1=0}^{k} \cdots \sum_{m_d=0}^{k} c_{m_1, m_2, \ldots, m_d} \, T_{m_1, \ldots, m_d}(x_1, x_2, \ldots, x_d).
\]

Thus, we have \( (k+1)^d \) parameters to estimate. While this estimation is feasible in low dimensions, the problem becomes significantly more difficult as the dimensionality \( d \) increases.

\subsection{Pairwise Decomposition for Dimensionality Reduction}

\begin{figure}[htbp]
    \centering
    \includegraphics[width=0.8\textwidth]{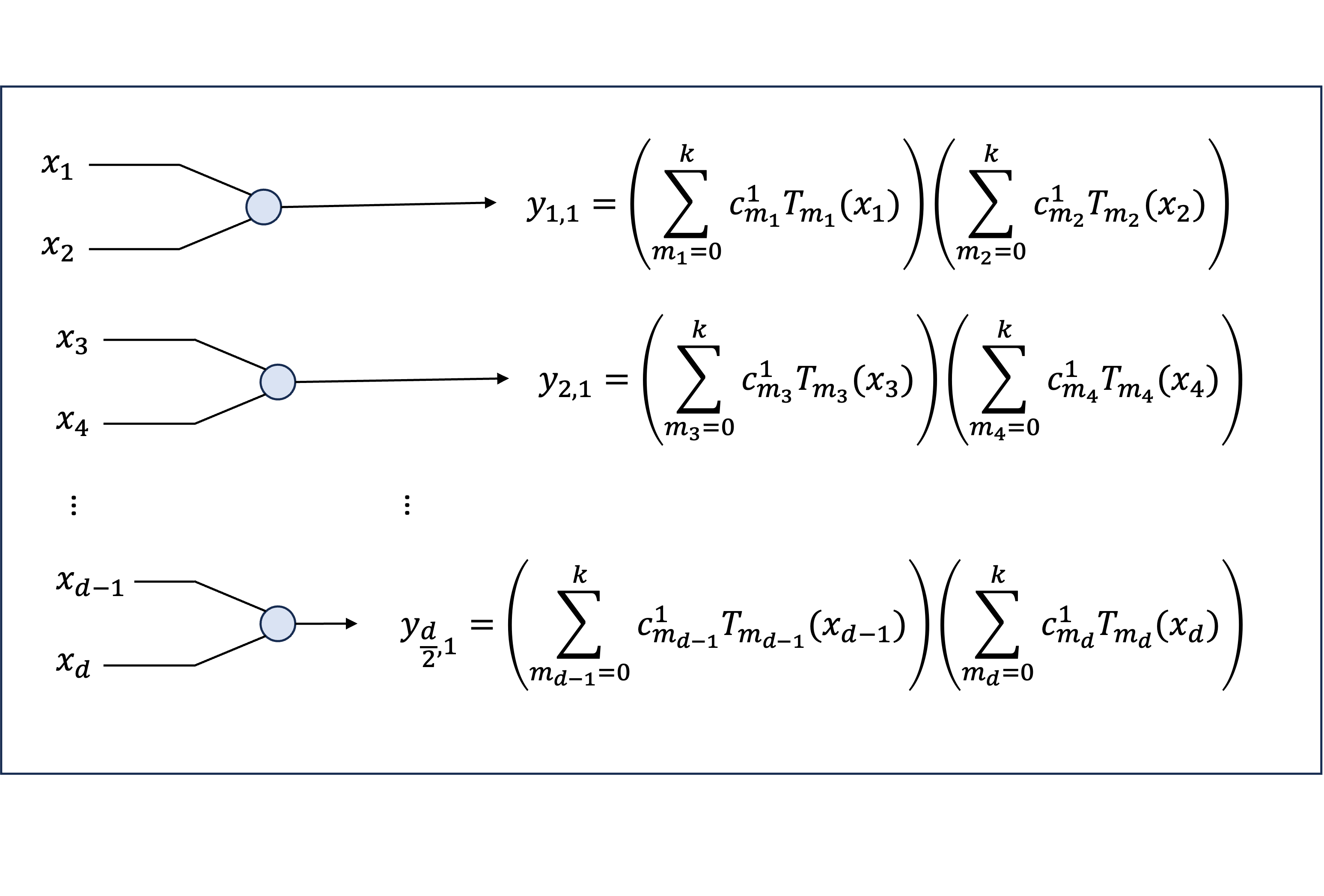}
    \caption{Initial layer of the pairwise decomposition, illustrating the pairwise approximation of variables \( x_1, x_2, \ldots, x_d \).}
    \label{fig:layer1}
\end{figure}

In high-dimensional cases, direct multivariate Chebyshev decomposition can lead to an excessive number of parameters. To address this, we propose a pairwise decomposition strategy to reduce the complexity of approximation. Suppose we have a function \( f(x_1, x_2, \ldots, x_d) \) and aim to classify or predict based on the features \( x_1, x_2, \ldots, x_d \). The goal is to approximate \( f(x_1, x_2, \ldots, x_d) \) through orthogonal decomposition using Chebyshev polynomials. Using Chebyshev decomposition, we approximate \( f(x_1, x_2, \ldots, x_d) \) as:

\[
f(x_1, \ldots, x_d) \approx \sum_{m_1=0}^{k} \cdots \sum_{m_d=0}^{k} c_{m_1, \ldots, m_d} T_{m_1}(x_1) \cdots T_{m_d}(x_d),
\]

where \( T_{m_i}(x_i) \) denotes the Chebyshev polynomial of degree \( m_i \) for each variable \( x_i \). This expression involves \( (k+1)^d \) parameters, which becomes computationally challenging as \( d \) increases. To reduce dimensionality, we assume pairwise combinations of variables, simplifying the expression by decomposing each pair:

\[
T_{m_1, m_2, \ldots, m_d}(x_1, x_2, \ldots, x_d) = \prod_{i=1}^{d} T_{m_i}(x_i).
\]

We then approximate \( f(x_1, \ldots, x_d) \) by summing over pairwise terms, effectively reducing the parameter space. We calculate intermediate outputs for each pairwise combination:

\[
y_{1,1} = \sum_{m_1=0}^{k} c_{m_1} T_{m_1}(x_1), \quad y_{2,1} = \sum_{m_2=0}^{k} c_{m_2} T_{m_2}(x_2),
\]

and so forth, until the final output is obtained by aggregating these pairwise decomposed terms. Through backpropagation, we estimate each coefficient \( c_j^i \), balancing approximation accuracy with computational feasibility. This method is particularly suitable for high-dimensional data where computational efficiency is critical.

\begin{figure}[htbp]
    \centering
    \includegraphics[width=0.8\textwidth]{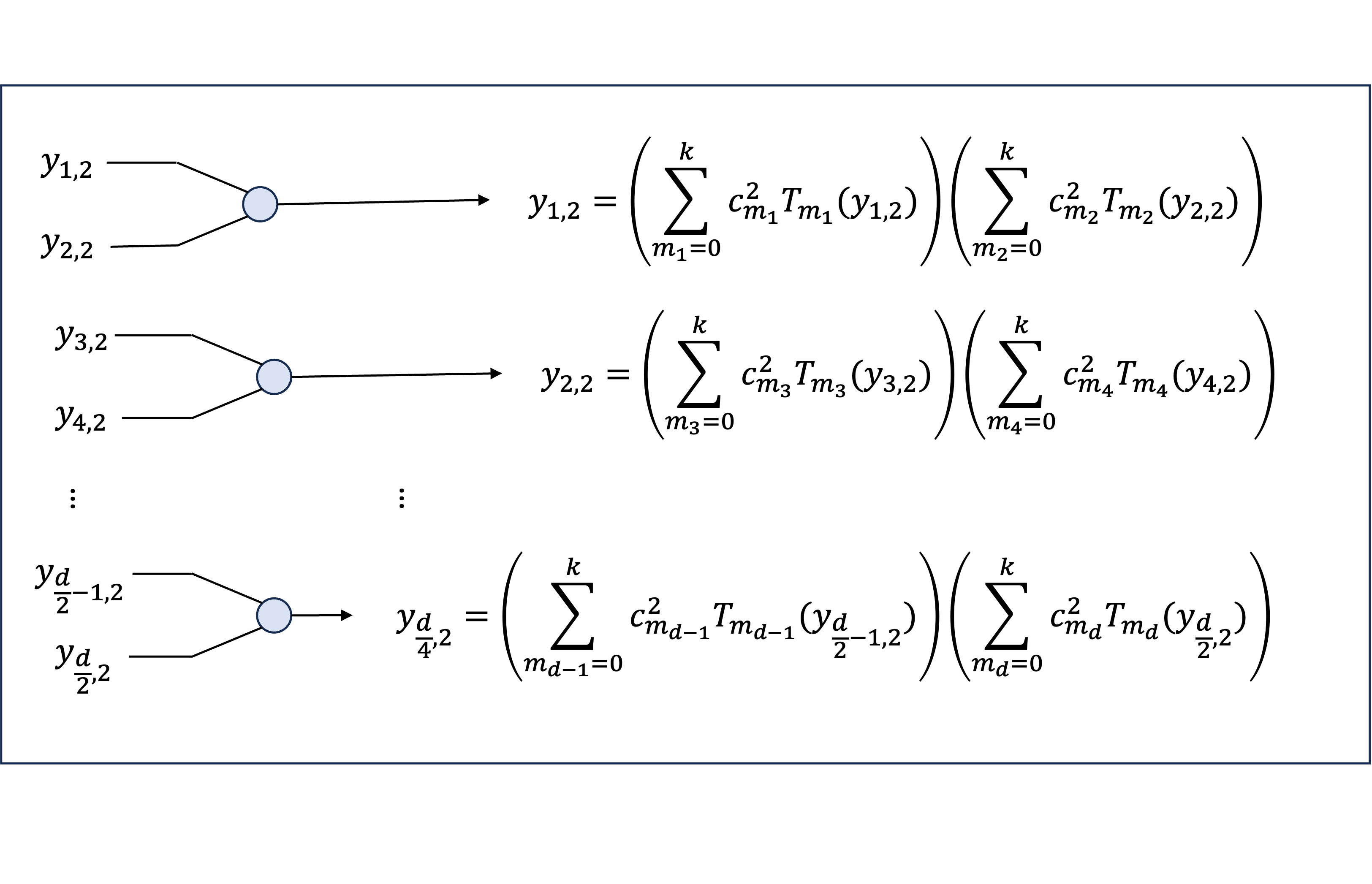}
    \caption{Intermediate layer showing further pairwise combinations to reduce parameter space complexity.}
    \label{fig:layer2}
\end{figure}

\begin{figure}[htbp]
    \centering
    \includegraphics[width=0.8\textwidth]{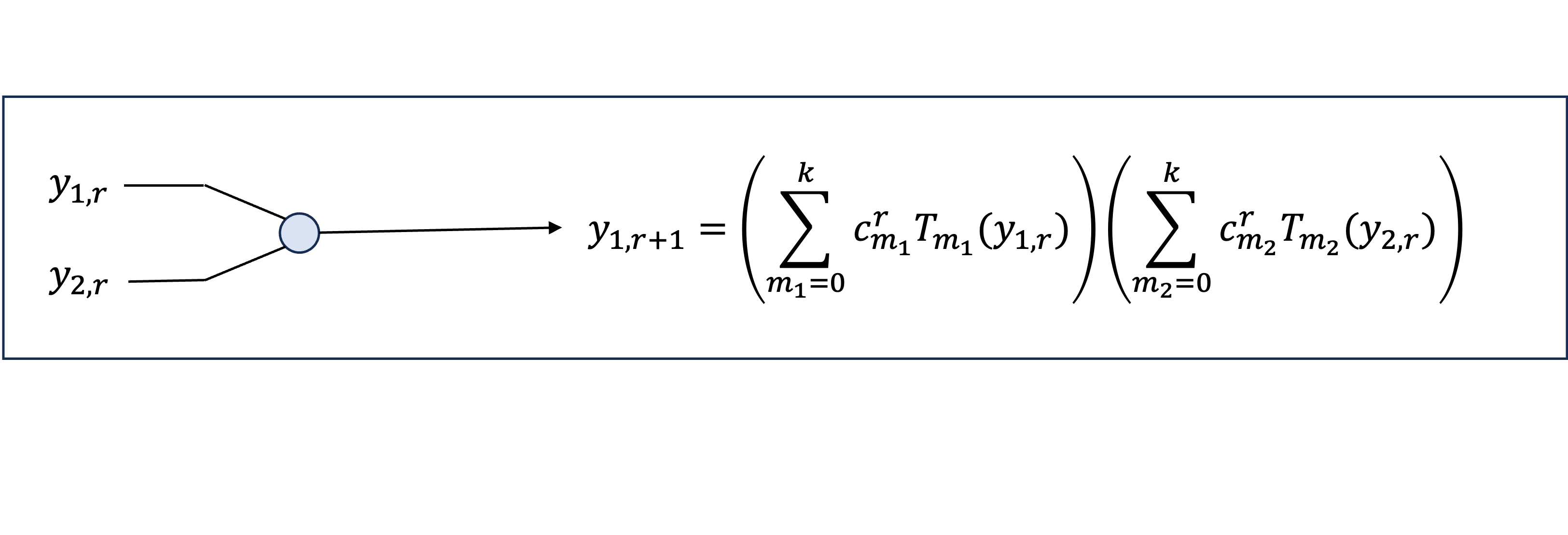}
    \caption{Final aggregation layer where pairwise approximations are combined to form the complete approximation of \( f(x_1, x_2, \ldots, x_d) \).}
    \label{fig:layer_n}
\end{figure}

In summary, Figures \ref{fig:layer1}, \ref{fig:layer2}, and \ref{fig:layer_n} illustrate the use of pairwise decomposition at each layer, which helps to handle the explosion of parameters in high-dimensional approximation. This approach leads to a final decomposition depth \( r+1 = \log_e(d) \), and backpropagation is used to estimate each coefficient \( c_j^i \), thereby enabling efficient and accurate high-dimensional function approximation.

\section{Conclusion}

This paper introduces an adaptive neural network model utilizing Chebyshev polynomials to achieve input-dependent weighting, emulating biological adaptability and enhancing the network's ability to capture complex, non-linear patterns. Empirical evaluation on 145 datasets from the PMLB benchmark reveals that the Chebyshev model outperforms traditional Multi-Layer Perceptrons (MLP) in 74\% of cases, achieving a mean accuracy of 84.13\% versus 80.87\% for MLP, with the highest accuracy gain reaching 26.83\%. Pruning further enhances efficiency, achieving up to 90\% compression without sacrificing performance. This adaptable framework shows promise in handling high-dimensional and out-of-distribution data, making it broadly applicable in areas with complex data dependencies, like healthcare and finance. Overall, this Chebyshev-based adaptive approach provides a flexible and efficient alternative to fixed-weight neural architectures, paving the way for future research into other orthogonal decompositions and high-dimensional adaptive models.

\appendix

\section{Complete Dataset description}
\begin{longtable}{p{3cm}p{2cm}p{3cm}p{3cm}}

\hline
\textbf{Dataset Name} & \textbf{Size} & \textbf{Number of Features} & \textbf{Number of Classes} \\
\hline
\endfirsthead

\hline
\textbf{Dataset Name} & \textbf{Size} & \textbf{Number of Features} & \textbf{Number of Classes} \\
\hline
\endhead

\hline
\endfoot

adult & 29305 & 14 & 2 \\
agaricus lepiota & 4887 & 22 & 2 \\
allbp & 2263 & 29 & 3 \\
allhyper & 2262 & 29 & 4 \\
allhypo & 2262 & 29 & 3 \\
allrep & 2263 & 29 & 4 \\
analcatdata aids & 30 & 4 & 2 \\
analcatdata asbestos & 49 & 3 & 2 \\
analcatdata authorship & 504 & 70 & 4 \\
analcatdata bankruptcy & 30 & 6 & 2 \\
analcatdata boxing1 & 72 & 3 & 2 \\
analcatdata boxing2 & 79 & 3 & 2 \\
analcatdata creditscore & 60 & 6 & 2 \\
analcatdata cyyoung8092 & 58 & 10 & 2 \\
analcatdata cyyoung9302 & 55 & 10 & 2 \\
analcatdata dmft & 478 & 4 & 6 \\
analcatdata fraud & 25 & 11 & 2 \\
analcatdata germangss & 240 & 5 & 4 \\
analcatdata happiness & 36 & 3 & 3 \\
analcatdata japansolvent & 31 & 9 & 2 \\
analcatdata lawsuit & 158 & 4 & 2 \\
ann thyroid & 4320 & 21 & 3 \\
appendicitis & 63 & 7 & 2 \\
australian & 414 & 14 & 2 \\
auto & 121 & 25 & 5 \\
backache & 108 & 32 & 2 \\
balance scale & 375 & 4 & 3 \\
biomed & 125 & 8 & 2 \\
breast & 419 & 10 & 2 \\
breast cancer & 171 & 9 & 2 \\
breast cancer wisconsin & 341 & 30 & 2 \\
breast w & 419 & 9 & 2 \\
buggyCrx & 414 & 15 & 2 \\
bupa & 207 & 5 & 2 \\
calendarDOW & 239 & 32 & 5 \\
car & 1036 & 6 & 4 \\
car evaluation & 1036 & 21 & 4 \\
cars & 235 & 8 & 3 \\
chess & 1917 & 36 & 2 \\
churn & 3000 & 20 & 2 \\
clean1 & 285 & 168 & 2 \\
clean2 & 3958 & 168 & 2 \\
cleve & 181 & 13 & 2 \\
cleveland & 181 & 13 & 5 \\
cleveland nominal & 181 & 7 & 5 \\
cmc & 883 & 9 & 3 \\
coil2000 & 5893 & 85 & 2 \\
colic & 220 & 22 & 2 \\
collins & 291 & 23 & 13 \\
connect 4 & 40534 & 42 & 3 \\
contraceptive & 883 & 9 & 3 \\
corral & 96 & 6 & 2 \\
credit a & 414 & 15 & 2 \\
credit g & 600 & 20 & 2 \\
crx & 414 & 15 & 2 \\
dermatology & 219 & 34 & 6 \\
diabetes & 460 & 8 & 2 \\
dis & 2263 & 29 & 2 \\
dna & 1911 & 180 & 3 \\
ecoli & 196 & 7 & 5 \\
fars & 60580 & 29 & 8 \\
flags & 106 & 43 & 5 \\
flare & 639 & 10 & 2 \\
GAMETES Epistasis 2 Way 1000atts 0.4H EDM 1 EDM 1 1 & 960 & 1000 & 2 \\
GAMETES Epistasis 2 Way 20atts 0.1H EDM 1 1 & 960 & 20 & 2 \\
GAMETES Epistasis 2 Way 20atts 0.4H EDM 1 1 & 960 & 20 & 2 \\
GAMETES Epistasis 3 Way 20atts 0.2H EDM 1 1 & 960 & 20 & 2 \\
GAMETES Heterogeneity 20atts 1600 Het 0.4 0.2 50 EDM 2 001 & 960 & 20 & 2 \\
GAMETES Heterogeneity 20atts 1600 Het 0.4 0.2 75 EDM 2 001 & 960 & 20 & 2 \\
german & 600 & 20 & 2 \\
glass & 123 & 9 & 5 \\
glass2 & 97 & 9 & 2 \\
haberman & 183 & 3 & 2 \\
hayes roth & 96 & 4 & 3 \\
heart c & 181 & 13 & 2 \\
heart h & 176 & 13 & 2 \\
heart statlog & 162 & 13 & 2 \\
hepatitis & 93 & 19 & 2 \\
Hill Valley with noise & 727 & 100 & 2 \\
Hill Valley without noise & 727 & 100 & 2 \\
horse colic & 220 & 22 & 2 \\
house votes 84 & 261 & 16 & 2 \\
hungarian & 176 & 13 & 2 \\
hypothyroid & 1897 & 25 & 2 \\
ionosphere & 210 & 34 & 2 \\
iris & 90 & 4 & 3 \\
irish & 300 & 5 & 2 \\
kr vs kp & 1917 & 36 & 2 \\
krkopt & 16833 & 6 & 18 \\
led24 & 1920 & 24 & 10 \\
led7 & 1920 & 7 & 10 \\
letter & 12000 & 16 & 26 \\
mfeat factors & 1200 & 216 & 10 \\
mfeat fourier & 1200 & 76 & 10 \\
mfeat karhunen & 1200 & 64 & 10 \\
mfeat morphological & 1200 & 6 & 10 \\
mfeat zernike & 1200 & 47 & 10 \\
mofn 3 7 10 & 794 & 10 & 2 \\
monk1 & 333 & 6 & 2 \\
monk2 & 360 & 6 & 2 \\
monk3 & 332 & 6 & 2 \\
movement libras & 216 & 90 & 15 \\
mushroom & 4874 & 22 & 2 \\
new thyroid & 129 & 5 & 3 \\
optdigits & 3372 & 64 & 10 \\
page blocks & 3283 & 10 & 5 \\
parity5+5 & 674 & 10 & 2 \\
penguins & 199 & 7 & 3 \\
phoneme & 3242 & 5 & 2 \\
pima & 460 & 8 & 2 \\
prnn crabs & 120 & 7 & 2 \\
prnn fglass & 123 & 9 & 5 \\
prnn synth & 150 & 2 & 2 \\
profb & 403 & 9 & 2 \\
ring & 4440 & 20 & 2 \\
saheart & 277 & 9 & 2 \\
satimage & 3861 & 36 & 6 \\
schizo & 204 & 14 & 3 \\
segmentation & 1386 & 19 & 7 \\
sleep & 63544 & 13 & 5 \\
solar flare 1 & 189 & 12 & 5 \\
solar flare 2 & 639 & 12 & 6 \\
sonar & 124 & 60 & 2 \\
soybean & 405 & 35 & 18 \\
spambase & 2760 & 57 & 2 \\
spect & 160 & 22 & 2 \\
spectf & 209 & 44 & 2 \\
splice & 1912 & 60 & 3 \\
tae & 90 & 5 & 3 \\
texture & 3300 & 40 & 11 \\
threeOf9 & 307 & 9 & 2 \\
tic tac toe & 574 & 9 & 2 \\
tokyo1 & 575 & 44 & 2 \\
twonorm & 4440 & 20 & 2 \\
vehicle & 507 & 18 & 4 \\
vote & 261 & 16 & 2 \\
vowel & 594 & 13 & 11 \\
waveform 21 & 3000 & 21 & 3 \\
waveform 40 & 3000 & 40 & 3 \\
wdbc & 341 & 30 & 2 \\
wine quality red & 959 & 11 & 6 \\
wine quality white & 2938 & 11 & 7 \\
wine recognition & 106 & 13 & 3 \\
xd6 & 583 & 9 & 2 \\
yeast & 887 & 8 & 9 \\
\label{tab:all_data_desc}
\end{longtable}

\section{Complete Results}

\begin{longtable}{p{3cm}p{3cm}p{3cm}p{3cm}}
\caption{Comparison of Chebyshev Adaptive Model and MLP Accuracy Across Datasets}
\label{tab:cheby_mlp_accuracy_comparison} \\
\hline
\textbf{Dataset Name} & \textbf{Chebyshev Adaptive Model Accuracy} & \textbf{MLP Accuracy} & \textbf{Difference (Chebyshev - MLP)} \\
\hline
\endfirsthead

\hline
\textbf{Dataset Name} & \textbf{Chebyshev Adaptive Model Accuracy} & \textbf{MLP Accuracy} & \textbf{Difference (Chebyshev - MLP)} \\
\hline
\endhead

\hline
\endfoot

auto & 75.61 & 48.78 & \textbf{26.829} \\
vowel & 86.869 & 60.606 & \textbf{26.263} \\
analcatdata fraud & 88.889 & 66.667 & \textbf{22.222} \\
soybean & 84.444 & 62.963 & \textbf{21.481} \\
tic tac toe & 98.438 & 79.167 & \textbf{19.271} \\
letter & 54.475 & 36.05 & \textbf{18.425} \\
movement libras & 58.333 & 43.056 & \textbf{15.278} \\
ring & 97.703 & 84.662 & \textbf{13.041} \\
car & 96.532 & 83.815 & \textbf{12.717} \\
hayes roth & 81.25 & 68.75 & \textbf{12.5} \\
analcatdata boxing2 & 88.889 & 77.778 & \textbf{11.111} \\
ecoli & 90.909 & 80.303 & \textbf{10.606} \\
analcatdata cyyoung9302 & 94.737 & 84.211 & \textbf{10.526} \\
analcatdata cyyoung8092 & 95 & 85 & \textbf{10}\\
analcatdata aids & 50 & 40 & 10 \\
sonar & 88.095 & 78.571 & 9.524 \\
analcatdata japansolvent & 90.909 & 81.818 & 9.091 \\
connect 4 & 79.677 & 70.634 & 9.044 \\
mfeat zernike & 79.75 & 71 & 8.75 \\
analcatdata boxing1 & 91.667 & 83.333 & 8.333 \\
analcatdata happiness & 58.333 & 50 & 8.333 \\
monk1 & 100 & 91.964 & 8.036 \\
calendarDOW & 51.25 & 43.75 & 7.5 \\
hungarian & 74.576 & 67.797 & 6.78 \\
solar flare 1 & 77.777 & 71.429 & 6.349 \\
collins & 96.907 & 90.722 & 6.186 \\
glass2 & 75.758 & 69.697 & 6.061 \\
analcatdata asbestos & 88.235 & 82.353 & 5.882 \\
backache & 94.444 & 88.889 & 5.556 \\
texture & 95.636 & 90.091 & 5.545 \\
krkopt & 39.273 & 34.141 & 5.132 \\
mfeat karhunen & 81.75 & 76.75 & 5 \\
Hill Valley with noise & 78.189 & 73.251 & 4.938 \\
cleveland & 62.295 & 57.377 & 4.918 \\
appendicitis & 100 & 95.455 & 4.545 \\
parity5+5 & 100 & 95.556 & 4.444 \\
splice & 90.439 & 86.05 & 4.389 \\
phoneme & 83.349 & 79.093 & 4.255 \\
mfeat factors & 92.5 & 88.25 & 4.25 \\
spect & 87.037 & 83.333 & 3.704 \\
breast cancer & 86.207 & 82.759 & 3.448 \\
cleve & 86.885 & 83.607 & 3.279 \\
heart c & 80.328 & 77.049 & 3.279 \\
GAMETES Epistasis 2 Way 20atts 0.4H EDM 1 1 & 75.938 & 72.813 & 3.125 \\
penguins & 98.507 & 95.522 & 2.985 \\
schizo & 61.765 & 58.824 & 2.941 \\
flags & 52.777 & 50 & 2.777 \\
mfeat morphological & 74.25 & 71.5 & 2.75 \\
GAMETES Epistasis 2 Way 1000atts 0.4H EDM 1 EDM 1 1 & 55 & 52.5 & 2.5 \\
GAMETES Heterogeneity 20atts 1600 Het 0.4 0.2 75 EDM 2 001 & 76.25 & 73.75 & 2.5 \\
analcatdata germangss & 37.5 & 35 & 2.5 \\
optdigits & 87.633 & 85.142 & 2.491 \\
glass & 73.171 & 70.732 & 2.439 \\
balance scale & 94.4 & 92 & 2.4 \\
satimage & 87.334 & 85.004 & 2.33 \\
saheart & 79.569 & 77.419 & 2.15 \\
led24 & 67.656 & 65.625 & 2.031 \\
yeast & 58.446 & 56.419 & 2.027 \\
mfeat fourier & 71.75 & 69.75 & 2 \\
prnn synth & 88 & 86 & 2 \\
credit g & 73.5 & 71.5 & 2 \\
segmentation & 96.97 & 95.022 & 1.948 \\
wine quality white & 55.204 & 53.265 & 1.939 \\
heart statlog & 90.741 & 88.889 & 1.852 \\
vehicle & 79.412 & 77.647 & 1.765 \\
monk2 & 100 & 98.347 & 1.653 \\
cleveland nominal & 60.656 & 59.016 & 1.639 \\
solar flare 2 & 77.77 & 76.19 & 1.58 \\
profb & 71.111 & 69.629 & 1.482 \\
crx & 91.304 & 89.855 & 1.449 \\
car evaluation & 97.399 & 95.954 & 1.445 \\
spectf & 82.857 & 81.429 & 1.429 \\
dna & 93.73 & 92.32 & 1.411 \\
colic & 90.541 & 89.189 & 1.351 \\
horse colic & 86.486 & 85.135 & 1.351 \\
cars & 82.278 & 81.013 & 1.266 \\
analcatdata dmft & 24.375 & 23.125 & 1.25 \\
GAMETES Heterogeneity 20atts 1600 Het 0.4 0.2 50 EDM 2 001 & 72.5 & 71.25 & 1.25 \\
GAMETES Epistasis 2 Way 20atts 0.1H EDM 1 1 & 68.125 & 66.875 & 1.25 \\
vote & 95.402 & 94.252 & 1.15 \\
sleep & 74.011 & 72.982 & 1.029 \\
threeOf9 & 99.029 & 98.058 & 0.971 \\
wdbc & 99.123 & 98.246 & 0.877 \\
Hill Valley without noise & 67.901 & 67.078 & 0.823 \\
allrep & 98.14 & 97.35 & 0.79 \\
credit a & 84.783 & 84.057 & 0.726 \\
australian & 89.13 & 88.406 & 0.725 \\
buggyCrx & 91.304 & 90.58 & 0.724 \\
breast w & 99.286 & 98.571 & 0.714 \\
contraceptive & 58.983 & 58.305 & 0.678 \\
allbp & 97.483 & 96.821 & 0.662 \\
pima & 79.221 & 78.571 & 0.649 \\
adult & 85.935 & 85.393 & 0.543 \\
tokyo1 & 93.75 & 93.229 & 0.521 \\
german & 74 & 73.5 & 0.5 \\
twonorm & 98.243 & 97.837 & 0.406 \\
churn & 93.2 & 92.8 & 0.4 \\
cmc & 57.627 & 57.288 & 0.339 \\
hypothyroid & 97.946 & 97.63 & 0.316 \\
GAMETES Epistasis 3 Way 20atts 0.2H EDM 1 1 & 55 & 54.688 & 0.313 \\
wine quality red & 58.125 & 57.813 & 0.313 \\
led7 & 68.75 & 68.438 & 0.313 \\
waveform 40 & 86.6 & 86.3 & 0.3 \\
allhypo & 95.093 & 94.828 & 0.265 \\
dis & 98.278 & 98.013 & 0.265 \\
fars & 78.028 & 77.795 & 0.233 \\
page blocks & 94.977 & 94.795 & 0.183 \\
ann thyroid & 98.888 & 98.75 & 0.138 \\
new thyroid & 97.674 & 97.674 & 0 \\
biomed & 92.857 & 92.857 & 0 \\
flare & 84.579 & 84.579 & 0 \\
bupa & 57.971 & 57.971 & 0 \\
house votes 84 & 98.851 & 98.851 & 0 \\
analcatdata lawsuit & 96.226 & 96.226 & 0 \\
mofn 3 7 10 & 100 & 100 & 0 \\
heart h & 83.051 & 83.051 & 0 \\
agaricus lepiota & 100 & 100 & 0 \\
analcatdata bankruptcy & 100 & 100 & 0 \\
ionosphere & 92.958 & 92.958 & 0 \\
irish & 100 & 100 & 0 \\
monk3 & 98.198 & 98.198 & 0 \\
kr vs kp & 99.531 & 99.531 & 0 \\
wine recognition & 100 & 100 & 0 \\
hepatitis & 90.323 & 90.323 & 0 \\
iris & 100 & 100 & 0 \\
clean2 & 100 & 100 & 0 \\
tae & 67.742 & 67.742 & 0 \\
prnn fglass & 70.732 & 70.732 & 0 \\
corral & 100 & 100 & 0 \\
xd6 & 100 & 100 & 0 \\
coil2000 & 93.893 & 93.893 & 0 \\
breast cancer wisconsin & 98.246 & 98.246 & 0 \\
chess & 99.531 & 99.531 & 0 \\
analcatdata creditscore & 95 & 95 & 0 \\
haberman & 77.419 & 77.419 & 0 \\
prnn crabs & 100 & 100 & 0 \\
allhyper & 98.278 & 98.278 & 0 \\
mushroom & 100 & 100 & 0 \\
breast & 100 & 100 & 0 \\
waveform 21 & 87.4 & 87.4 & 0 \\
spambase & 94.354 & 94.354 & 0 \\
diabetes & 80.519 & 80.519 & 0 \\
analcatdata authorship & 99.408 & 99.408 & 0 \\
clean1 & 100 & 100 & 0 \\
dermatology & 97.297 & 97.297 & 0 \\
\end{longtable}

\section{Performance of Pruned Chebyshev model}

Comparison of pruned Chebyshev adaptive neural networks with standard MLPs across a variety of datasets. Table \ref{tab:pruned_chebyshev_gains_complete} showcases accuracy gains achieved through pruning, with datasets organized by decreasing levels of compression. Pruning was specifically applied to networks where the unpruned Chebyshev model matched the accuracy of the MLP, leading to notable improvements in accuracy and significant model compression, reaching up to ~90\% reduction in model size. The columns present dataset names, MLP accuracy, pruned Chebyshev model accuracy, and the achieved compression percentage.

\begin{table}[htbp]
\centering
\begin{tabular}{p{3cm}p{2cm}p{3cm}p{3cm}}
\hline
\textbf{Dataset} & \textbf{MLP Accuracy} & \textbf{Chebyshev Adaptive Model Accuracy (Pruned)} & \textbf{Compression} \\
\hline
wdbc & 98.246 & 99.123 & 89.2 \\
\hline
credit a & 84.057 & 84.783 & 89.1 \\
\hline
tokyo1 & 93.229 & 93.75 & 86.3 \\
\hline
breast & 100 & 100 & 85.8 \\
\hline
pima & 78.571 & 79.221 & 82.9 \\
\hline
cmc & 57.288 & 57.627 & 80.3 \\
\hline
cleveland nominal & 59.016 & 60.656 & 79.7 \\
\hline
GAMETES Heterogeneity 20atts 1600 Het 0.4 0.2 75 EDM 2 001 & 73.75 & 76.25 & 77.5 \\
\hline
GAMETES Epistasis 2 Way 20atts 0.1H EDM 1 1 & 66.875 & 68.125 & 75.3 \\
\hline
contraceptive & 58.305 & 58.983 & 74.2 \\
\hline
buggyCrx & 90.58 & 91.304 & 74 \\
\hline
heart statlog & 88.889 & 90.741 & 73.8 \\
\hline
profb & 69.629 & 71.111 & 69.5 \\
\hline
cleve & 83.607 & 86.885 & 66.7 \\
\hline
flags & 50 & 52.777 & 66.6 \\
\hline
colic & 89.189 & 90.541 & 61 \\
\hline
GAMETES Epistasis 2 Way 1000atts 0.4H EDM 1 EDM 1 1 & 52.5 & 55 & 57.6 \\
\hline
GAMETES Heterogeneity 20atts 1600 Het 0.4 0.2 50 EDM 2 001 & 71.25 & 72.5 & 56.7 \\
\hline
Hill Valley with noise & 73.251 & 78.189 & 55.2 \\
\hline
saheart & 77.419 & 79.569 & 52 \\
\hline
analcatdata asbestos & 82.353 & 88.235 & 49 \\
\hline
german & 73.5 & 74 & 48.1 \\
\hline
breast cancer & 82.759 & 86.207 & 46 \\
\hline
\end{tabular}
\caption{Performance of pruned Chebyshev adaptive neural networks compared to standard MLPs across various datasets. The table highlights improvements in accuracy achieved by pruning, with the datasets arranged in decreasing order of model compression. For datasets where the unpruned Chebyshev model matched MLP performance, pruning not only increased accuracy but also resulted in substantial model compression, achieving up to ~90\% reduction in model size while enhancing accuracy. The table columns display the dataset names, MLP accuracy, pruned Chebyshev model accuracy, and the compression percentage.}
\label{tab:pruned_chebyshev_gains_complete}
\end{table}

\section{Distribution of Weight for Different Datasets}

In this section, we examine the adaptive value of the neural network parameters with changing inputs. After training a Chebyshev adaptive model, we pass input values within the range of -1 to 1 and observe the corresponding adaptive weight values. These values are then plotted to illustrate how the weights respond to varying inputs, enhancing the model's capacity for tasks such as classification. The curve shapes in these plots indicate that the adaptive weights effectively respond to input variations, enabling improved performance in downstream tasks.

\begin{figure}[htbp]
    \centering
    \includegraphics[width=0.5\textwidth]{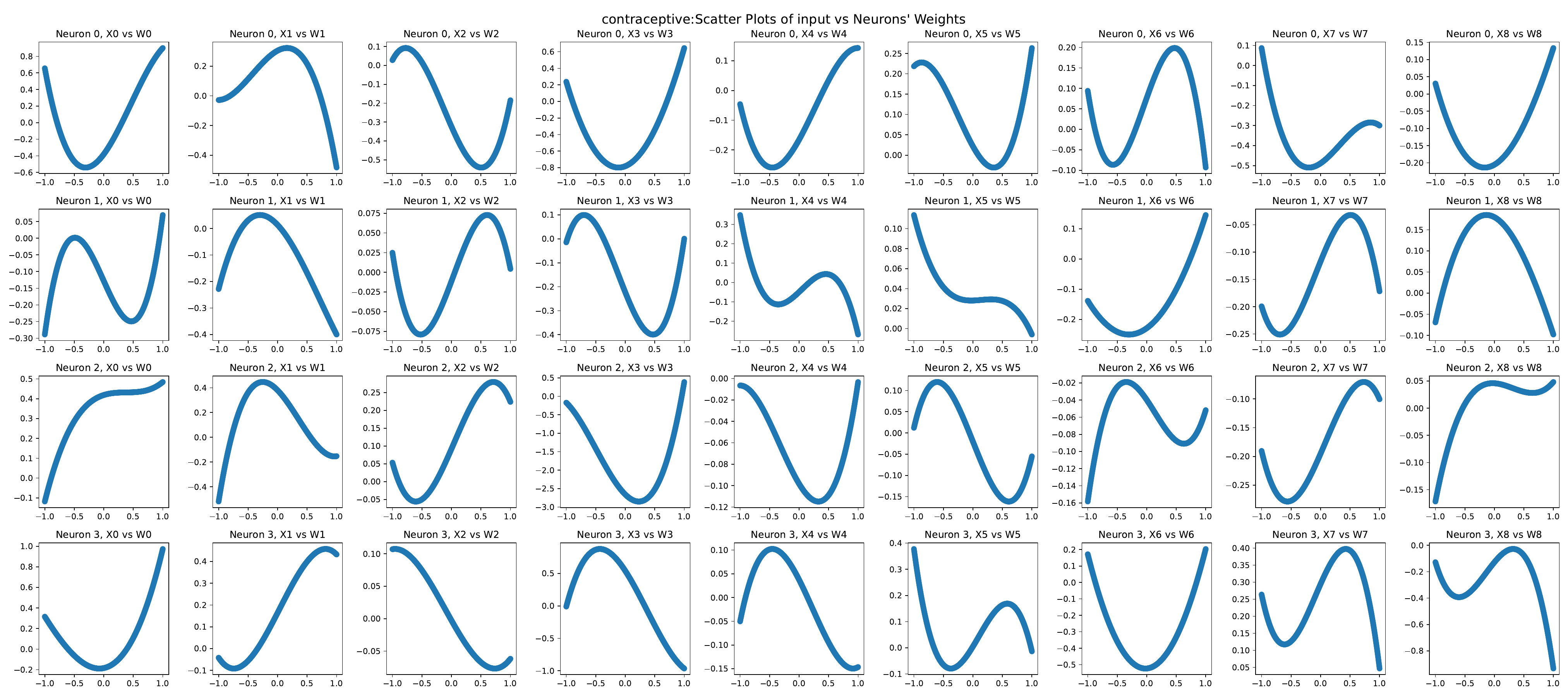}
    \caption{Adaptive weight distribution for the \textit{contraceptive} dataset, showing the weight's smooth variation with input values, reflecting its adaptive behavior.}
    \label{fig:contraceptive_weights}
\end{figure}

\begin{figure}[htbp]
    \centering
    \includegraphics[width=0.5\textwidth]{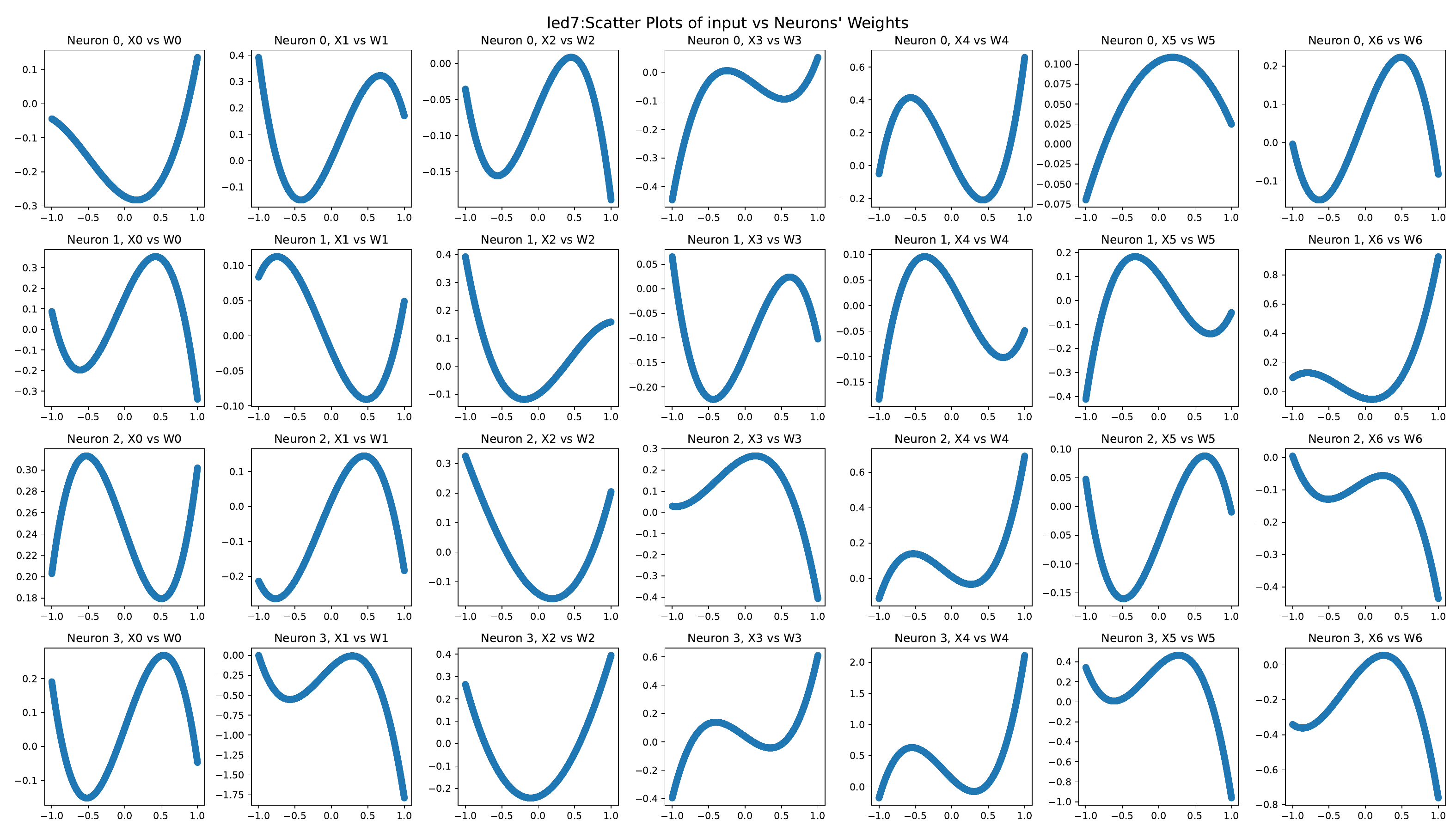}
    \caption{Adaptive weight distribution for the \textit{led7} dataset, illustrating the responsive nature of weights to input changes within the Chebyshev adaptive model.}
    \label{fig:led7_weights}
\end{figure}

\begin{figure}[htbp]
    \centering
    \includegraphics[width=0.5\textwidth]{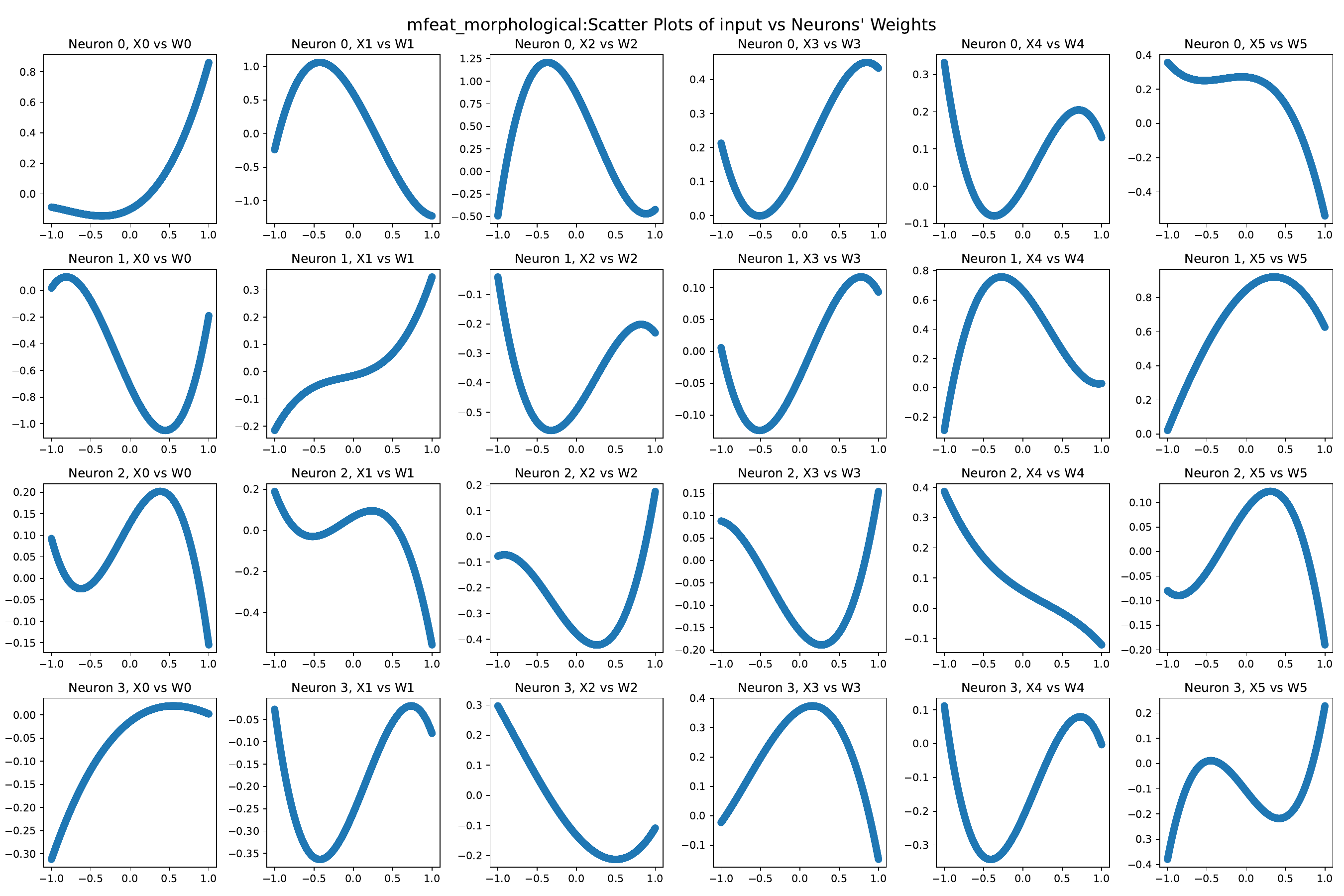}
    \caption{Adaptive weight distribution for the \textit{mfeat\_morphological} dataset, demonstrating the input-dependent weight adaptation in the model.}
    \label{fig:mfeat_morphological_weights}
\end{figure}

Figures \ref{fig:contraceptive_weights}, \ref{fig:led7_weights}, and \ref{fig:mfeat_morphological_weights} depict the adaptive weight distribution for the \textit{contraceptive}, \textit{led7}, and \textit{mfeat\_morphological} datasets, respectively. These figures highlight the smooth response of weights to the input variations, supporting the model's improved classification capabilities and flexibility.

\section{Decision Boundaries for Different Datasets}

In addition to the decision boundaries shown in the main paper, we provide a comparison of decision boundaries between the MLP and Chebyshev adaptive neural networks across various datasets. We include both binary and multiclass datasets to illustrate the Chebyshev model's superior ability to capture complex, non-linear boundaries. This is evident in the visualizations where the Chebyshev adaptive neural network significantly outperforms the MLP, particularly in datasets with intricate class distributions. The following figures illustrate these boundaries for each dataset.

\begin{figure}[htbp]
    \centering
    \includegraphics[width=0.8\textwidth]{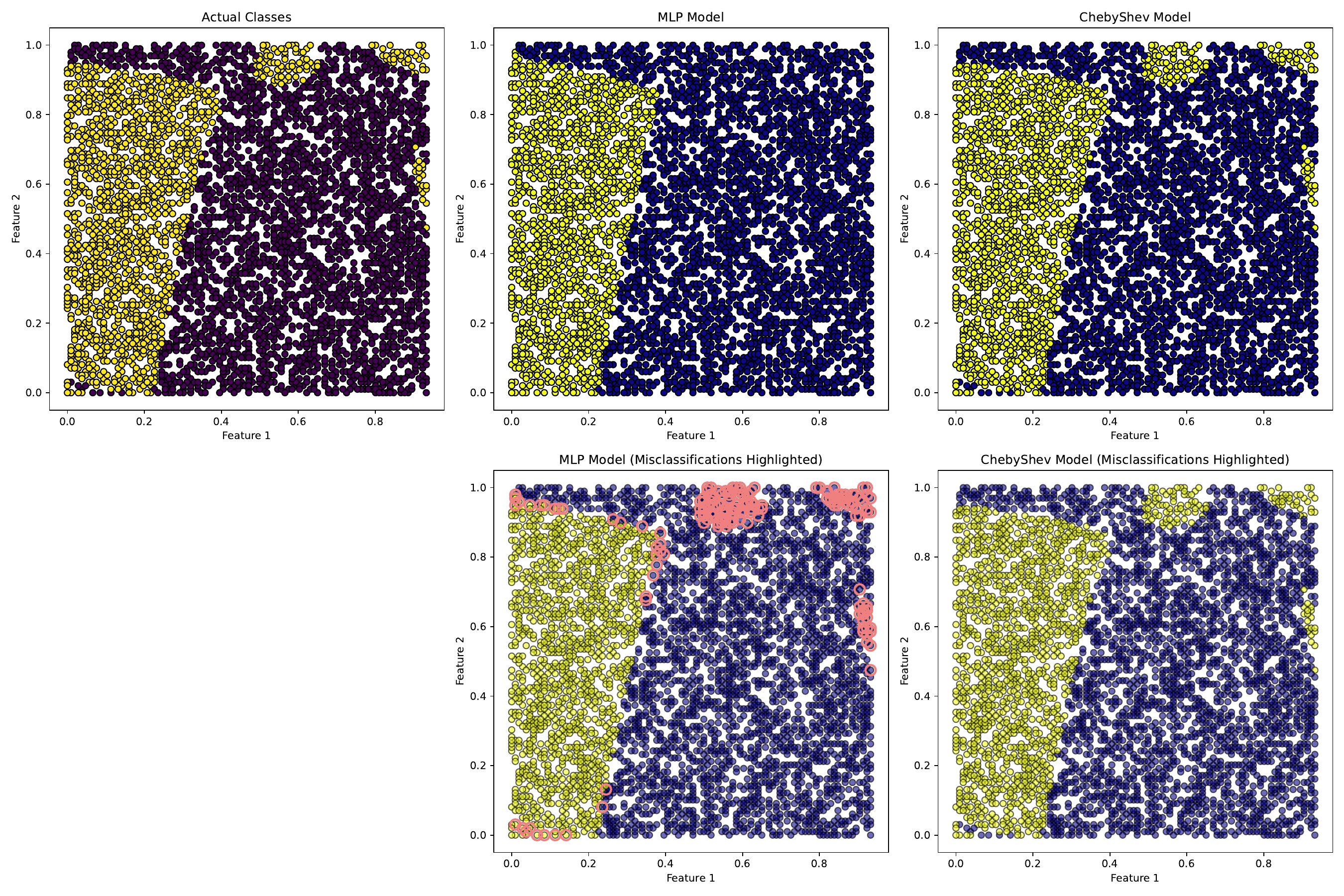}
    \caption{Comparison of decision boundaries on the "Tokyo1" dataset. The first plot in the top row shows the actual class distribution. The second plot shows the decision boundaries learned by the MLP, followed by those of the Chebyshev model. Below each plot, the misclassifications by each model are highlighted in pink.}
    \label{fig:tokyo1_boundaries}
\end{figure}

\begin{figure}[htbp]
    \centering
    \includegraphics[width=0.8\textwidth]{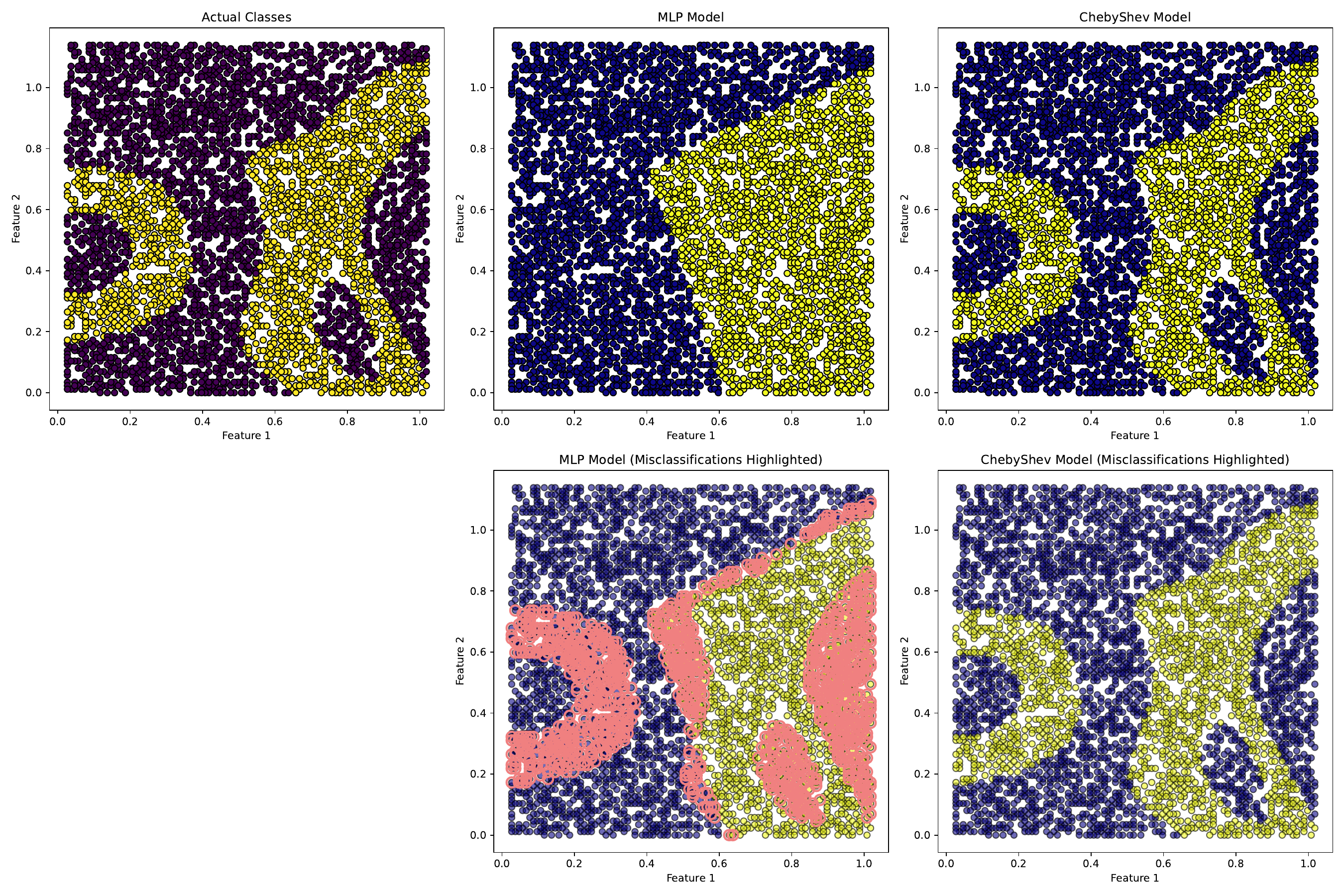}
    \caption{Comparison of decision boundaries on the "SAHeart" dataset. The top row displays the actual class distribution, followed by the decision boundaries of the MLP and Chebyshev models. The bottom row shows misclassifications for each model, with pink regions indicating errors.}
    \label{fig:saheart_boundaries}
\end{figure}

\begin{figure}[htbp]
    \centering
    \includegraphics[width=0.8\textwidth]{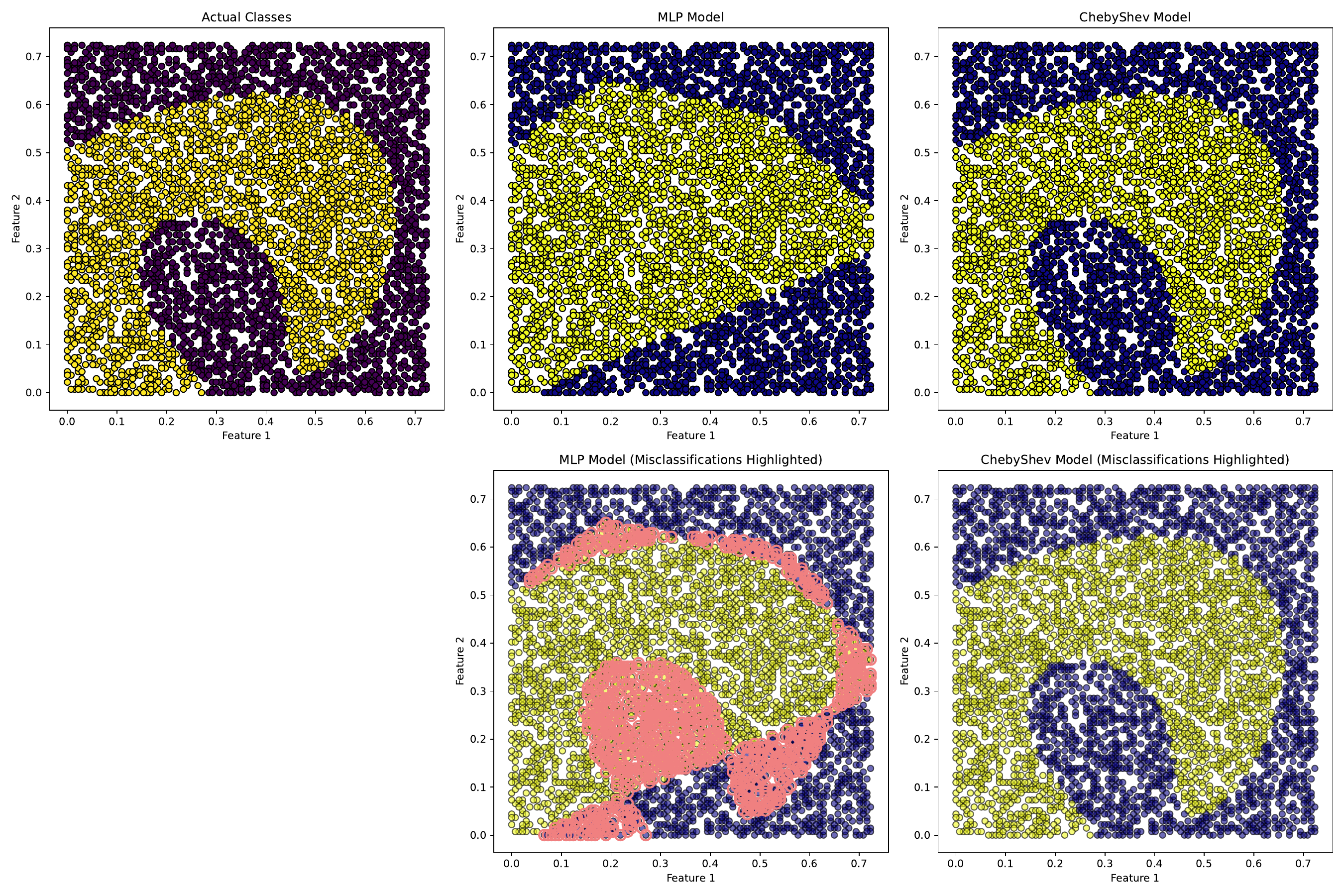}
    \caption{Decision boundaries comparison on the "Hill Valley without Noise" dataset. The first plot shows the actual classes, followed by the MLP and Chebyshev model boundaries. Misclassifications are highlighted in the bottom row, with pink areas indicating errors for each model.}
    \label{fig:hill_valley_boundaries}
\end{figure}

\begin{figure}[htbp]
    \centering
    \includegraphics[width=0.8\textwidth]{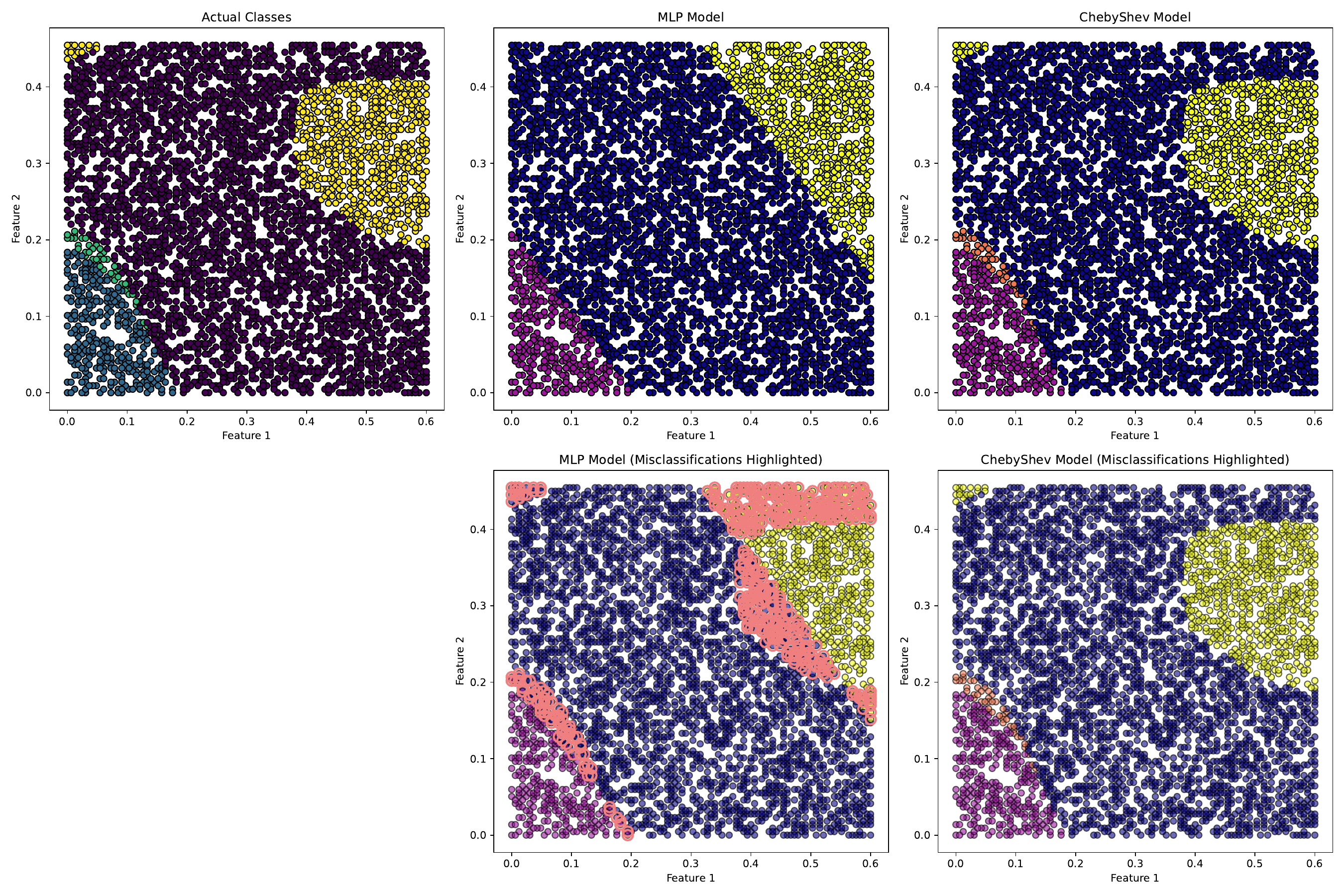}
    \caption{Decision boundaries on the "Flags" dataset. The initial plot displays the actual class distribution, followed by MLP and Chebyshev model boundaries. Misclassifications are presented in the bottom row for each model, with pink regions highlighting errors.}
    \label{fig:flags_boundaries}
\end{figure}

\begin{figure}[htbp]
    \centering
    \includegraphics[width=0.8\textwidth]{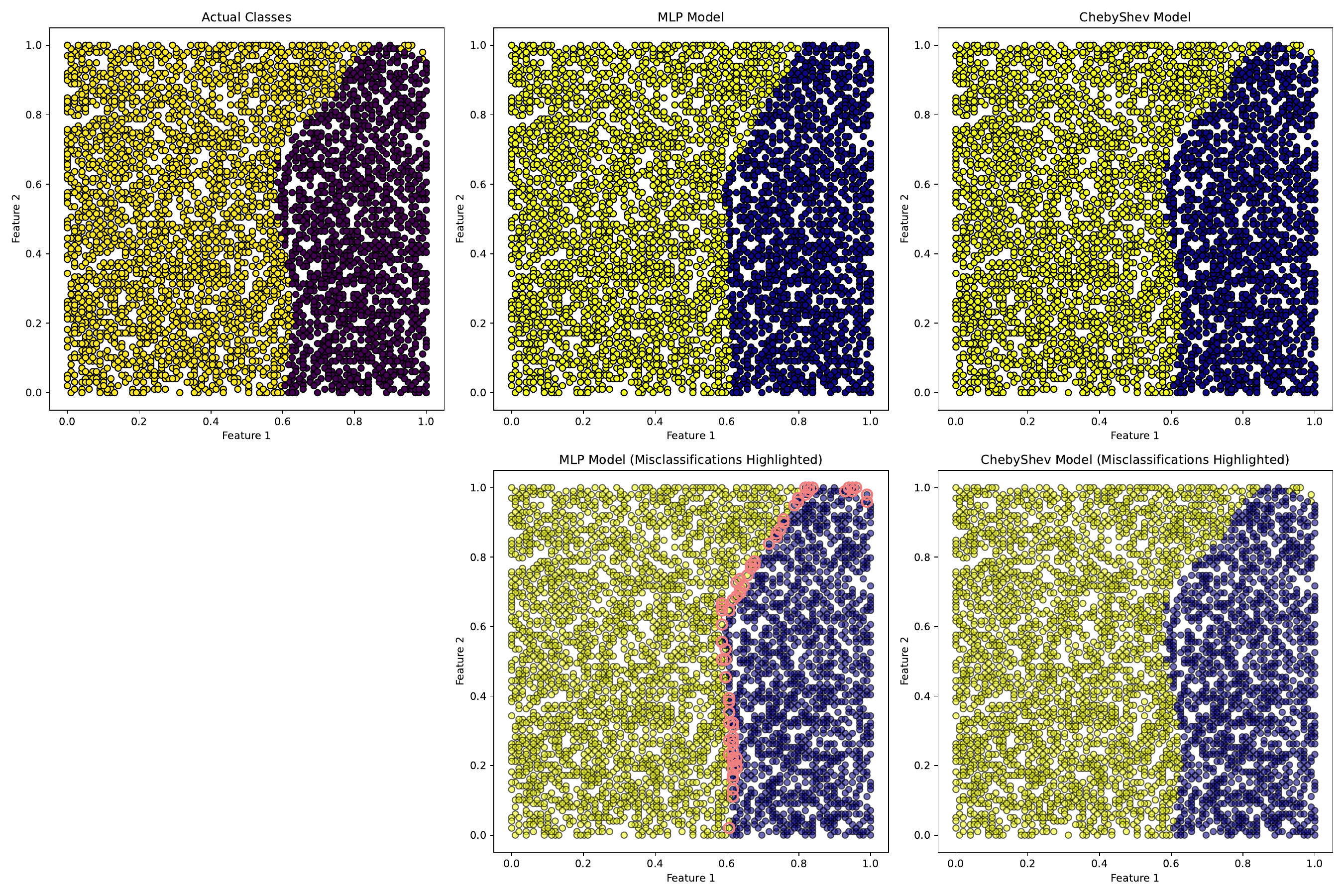}
    \caption{Decision boundary comparison on the "CMC" dataset. The top row contains the actual distribution and the decision boundaries of MLP and Chebyshev models. The misclassifications are shown in the bottom row for each model, with pink areas indicating errors.}
    \label{fig:cmc_boundaries}
\end{figure}

\begin{figure}[htbp]
    \centering
    \includegraphics[width=0.8\textwidth]{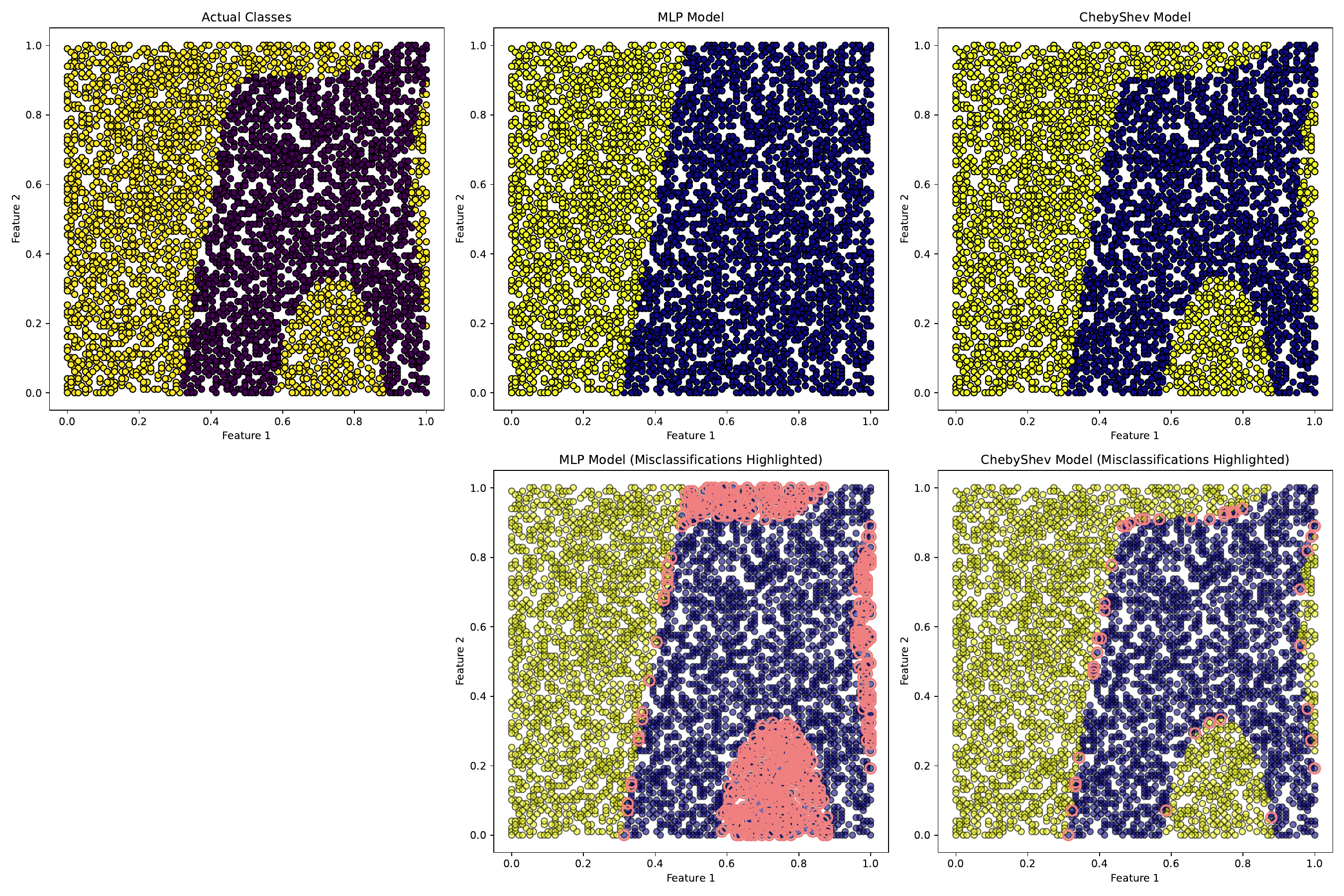}
    \caption{Comparison of decision boundaries on the "Clean2" dataset. The first plot shows the actual classes, followed by the boundaries learned by MLP and Chebyshev models. Misclassifications are highlighted in pink for each model in the bottom row.}
    \label{fig:clean2_boundaries}
\end{figure}

\begin{figure}[htbp]
    \centering
    \includegraphics[width=0.8\textwidth]{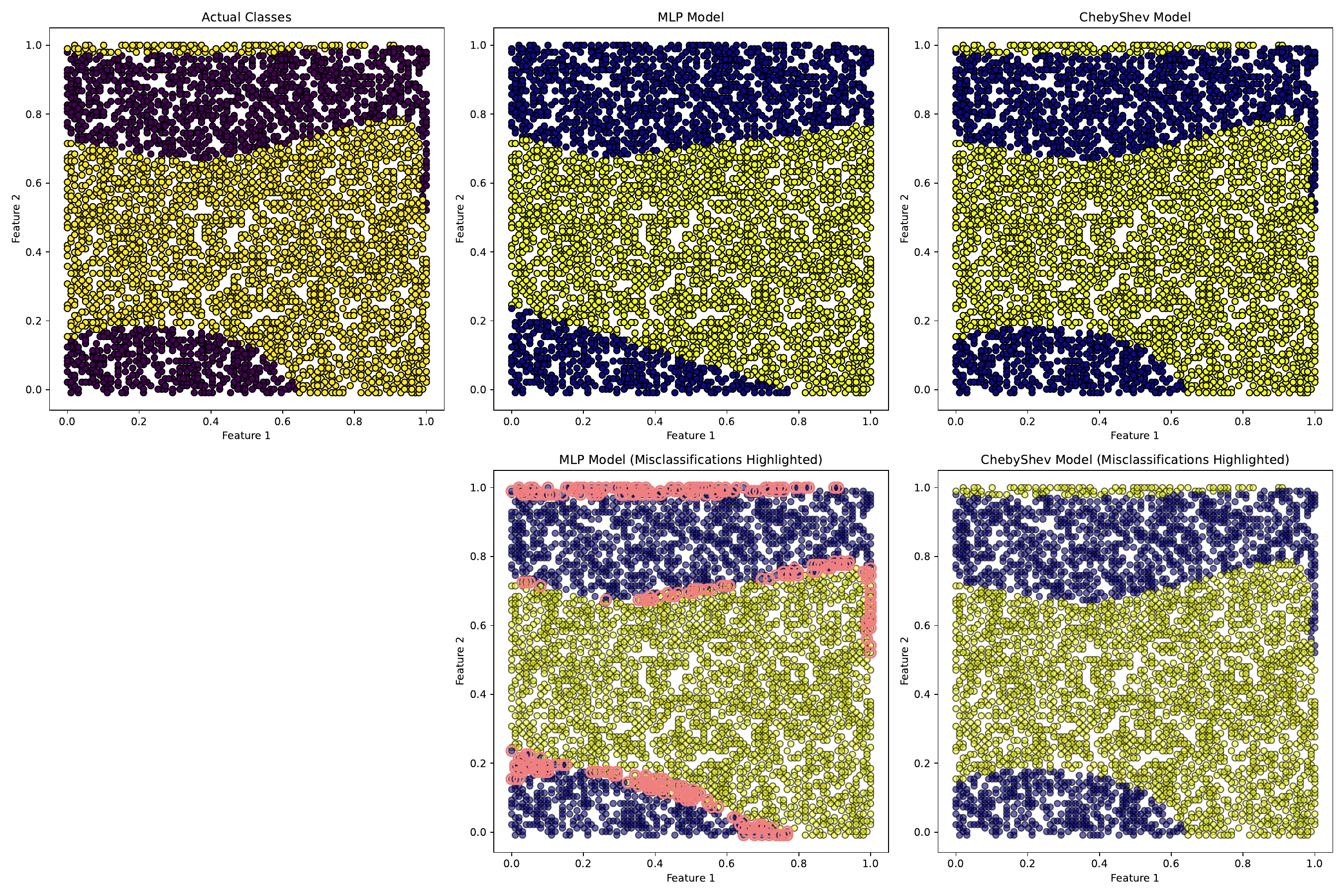}
    \caption{Decision boundaries for the "BuggyCrx" dataset. The top row displays the actual class distribution, followed by the decision boundaries of the MLP and Chebyshev models. Misclassifications are shown in pink in the bottom row for each model.}
    \label{fig:buggyCrx_boundaries}
\end{figure}

These figures collectively demonstrate the Chebyshev model's superiority in developing precise decision boundaries, especially in datasets with complex, non-linear separations between classes.

\bibliographystyle{elsarticle-harv}
\bibliography{elsarticle-template-num-names}





\end{document}